\documentclass[lettersize,journal]{IEEEtran}
\usepackage{amsmath,amsfonts}
\usepackage{algorithmic}
\usepackage{array}
\usepackage[caption=false,font=normalsize,labelfont=sf,textfont=sf]{subfig}
\usepackage{textcomp}
\usepackage{stfloats}
\usepackage{url}
\usepackage{verbatim}
\usepackage{graphicx}

\usepackage{amsmath} % assumes amsmath package installed
\usepackage{amssymb}  % assumes amsmath package installed

\usepackage{caption}
\usepackage{color}

\def\revised_color{red}

\newcommand{\reffig}[1]{{Fig. \ref{#1}}}

\newcommand{\refsect}[1]{{Section \ref{#1}}}
\newcommand{\refeq}[1]{{(\ref{#1})}}

\def\eq_color{black}

\def\BibTeX{{\rm B\kern-.05em{\sc i\kern-.025em b}\kern-.08em
    T\kern-.1667em\lower.7ex\hbox{E}\kern-.125emX}}
\usepackage{balance}
\begin{document}
\title{Adaptive-twist Soft Finger Mechanism for Grasping by Wrapping}
\author{Hiroki Ishikawa, Kyosuke Ishibashi and Ko Yamamoto, \it{Member, IEEE}% <-this % stops a space
\thanks{Manuscript received: March 24, 2025; Revised: July 29, 2025; Accepted: Sep. 30, 2025. This paper was recommended for publication by Editor Yong-Lae Park upon evaluation of the Associate Editor and Reviewers’ comments.}
\thanks{The authors are with the Department of Mechano-
Informatics, the University of Tokyo, Tokyo 113-8656, Japan (e-mail:
\{ishikawa-hiroki, ishibashi-kyosuke, yamamoto.ko\}@ynl.t.u-tokyo.ac.jp).}% <-this % stops a space
\thanks{*This research was supported by JSPS KAKENHI Grant Number 21H01282.}% <-this % stops a space
\thanks{Digital Objec Identifier (DOI): 10.1109/LRA.2025.3621979}
}

\markboth{IEEE ROBOTICS AND AUTOMATION LETTERS. PREPRINT VERSION. ACCEPTED SEPTEMBER, 2025 }
{Ishikawa et al.: Adaptive-twist Soft Finger Mechanism for Grasping by Wrapping}

\maketitle

\begin{abstract}
This paper presents a soft robot finger capable of adaptive-twist deformation to grasp objects by wrapping them.
For a soft hand to grasp and pick-up one object from densely contained multiple objects, a soft finger requires the \textbf{\textit{adaptive-twist}} deformation function in both in-plane and out-of-plane directions. 
The function allows the finger to be inserted deeply into a limited gap among objects.
Once inserted, the soft finger requires appropriate control of grasping force normal to contact surface, thereby maintaining the twisted deformation.
In this paper, we refer to this type of grasping as \textbf{\textit{grasping by wrapping}}.
To achieve these two functions by a single actuation source, we propose a variable stiffness mechanism that can adaptively change the stiffness as the pressure is higher.
We conduct a finite element analysis (FEA) on the proposed mechanism and determine its design parameter based on the FEA result.
Using the developed soft finger, we report basic experimental results and demonstrations on grasping various objects.
\end{abstract}


\section{Introduction\label{sect:intro}}
There is great demand for task automation across industries, especially in the agricultural and food industries, because of constantly shrinking work-age population.
In a vegetable factory, for example, the task of picking and transportation of large and heavy vegetables, such as cabbage, is manually performed, which places a heavy burden on workers.
Soft robotic hands \cite{rus2015design}
%\cite{rus2015design,soft_robotic_hand_review_shintake2018soft}
have many potential applications owing to their adaptability, including in
 food manipulation \cite{yamanaka2020development,food_handling_wang2017prestressed}.
Lightweight food items including vegetables require soft and gentle grasping and manipulation, which are the key features of a soft robotic hand.
There are several actuation methods for soft robots, including wire-driven systems \cite{tendon-calisti2011octopus,tendon_driven_gunderman2022tendon,shape_memory_alloy_lee2019long}, the use of shape-memory alloys \cite{SMA-laschi2012soft,shape_memory_alloy_lee2019long}, pneumatics \cite{jabara-mosadegh2014pneumatic,yap2016high} and hydraulics \cite{azami2023development}.
In particular, hydraulic actuation allows for higher pressure and stiffness, which is suitable for manipulating a large and heavy vegetables \cite{azami2023development}.

%%%%%%%%%%%%%%%%%%%%%%%%%%%%%%%%%%%%%%%%%%%%%%%%
\begin{figure}[t]
    \centering
    \includegraphics[width=1.0\hsize]{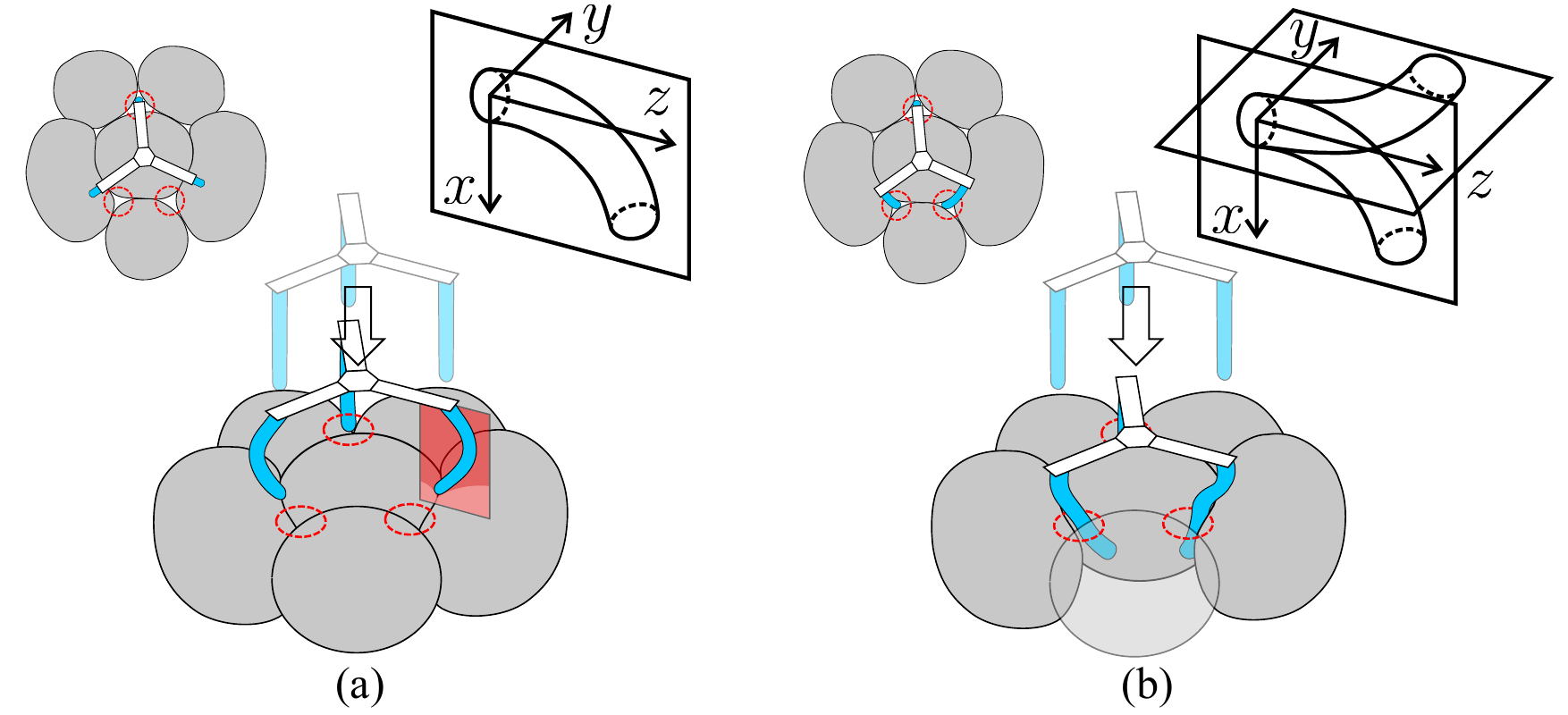}
  \caption{(a) A soft finger that only allows {\it in-plane} deformation. As shown, it is sometimes difficult to insert all fingers into unequal gap spaces among densely contained objects. (b) A soft finger that also allows {\it out-of-plane} directions can be easily inserted into gap spaces by {\it adaptively-twisting} deformation.}
  \label{fig:adaptive_twist_concept}
\end{figure}
%%%%%%%%%%%%%%%%%%%%%%%%%%%%%%%%%%%%%%%%%%%%%%%%
%%%%%%%%%%%%%%%%%%%%%%%%%%%%%%%%%%%%%%%%%%%%%%%%
\begin{figure}[t]
    \centering
    \subfloat[\label{fig:wrapping_grasp_concept_normal_direction}]{
        \includegraphics[width=0.15\hsize]{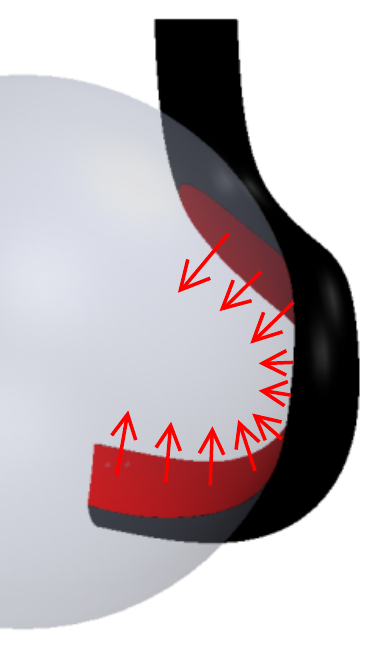}
    }
    \subfloat[\label{fig:wrapping_grasp_concept_tangent_direction}]{
        \includegraphics[width=0.35\hsize]{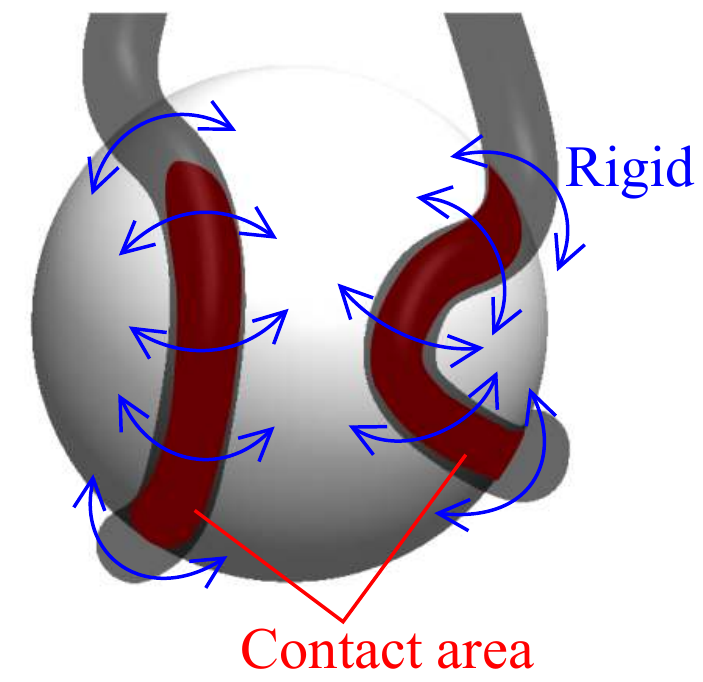}
    }
    \subfloat[\label{fig:grab_various_objects_ball}]{
        \includegraphics[width=0.35\hsize]{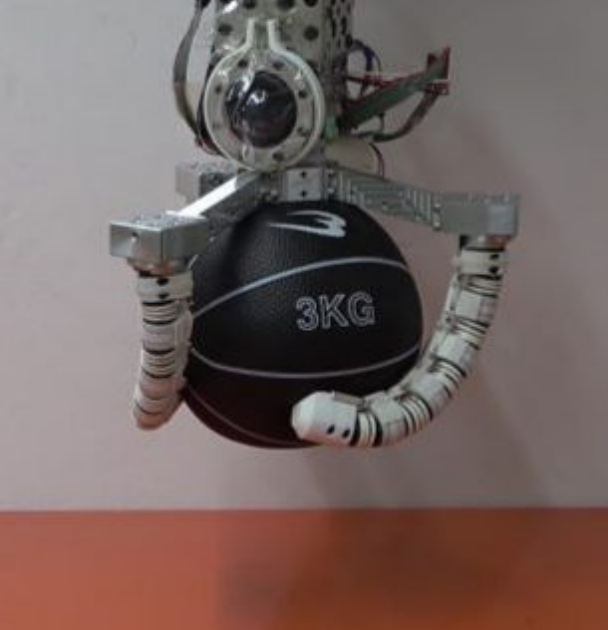}
    }
  \caption{Concept of {\it grasping by wrapping} and the developed soft finger.}
  \label{fig:wrapping_grasp_concept}
\end{figure}
%%%%%%%%%%%%%%%%%%%%%%%%%%%%%%%%%%%%%%%%%%%%%%%%
Figure \ref{fig:adaptive_twist_concept} shows a soft hand grasping and picking up an object from densely contained multiple objects.
One example is the picking task of large and heavy cabbages that are densely contained in a box, as seen in a vegetable factory.
A long and thin finger is suitable for a soft hand performing this task because it can be deeply inserted into limited gaps among objects.
However, the randomness of number, position and size of gaps complicates the task.
For example, \reffig{fig:adaptive_twist_concept} shows a case in which a soft hand has three fingers equally aligned whereas there are five gaps that are not equally spaced. 
To adapt to such a situation, a soft robot requires the following functions:
\begin{enumerate}
    \item 
    A finger should be able to adapt to the location of a gap.
    A soft finger with {\it in-plane} directional softness (in the $xz$ plane shown in \reffig{fig:adaptive_twist_concept}(a)) has low adaptability for unequally-spaced gaps.
    However, a finger with at least two-directional softness as shown in \reffig{fig:adaptive_twist_concept}(b), which is usually referred to as {\it out-of-plane} motion, can adapt to this situation by {\it twisting} deformation in the three-dimensional space.
    \item Once inserted, the soft finger requires appropriate control of grasping force normal to contact surface, as shown in \reffig{fig:wrapping_grasp_concept_normal_direction}, while maintaining the twisted deformation.
    In this study, we refer to this type of grasping as {\it grasping by wrapping}.
    If a soft finger has a flexibility in directions tangential to the contact surface, the actuation force is not directly converted to the grasping force in the normal direction.
    Therefore, a soft finger should be rigid in the tangential direction while grasping an object, as shown in \reffig{fig:wrapping_grasp_concept_tangent_direction}.
\end{enumerate}

According to the basic principle of soft robot actuation, deformation in the desired directions is extracted from the expansion of the soft material with the other directions constrained \cite{suzumori1991flexible}.
Usually, the number of deformable directions is equal to that of the actuation sources, which ensures that the actuation force is directly transformed into deformation.
Therefore, the aforementioned two functions can be achieved by multiple actuation sources, which usually requires a complicated structure consisting of multiple flow paths or chambers corresponding to the number of actuation sources \cite{multidirectional-pneumatic-zhang2019design}.
However, the control algorithm for grasping by wrapping is complicated.
%Also, when we select NBR as a material with high strength, it is difficult to fabricate such a complicate shape.
In particular, it is difficult to fabricate such a complicated shape using materials with a high tensile strength, such as nitrile-butadiene rubber (NBR), when using hydraulics to generate a pressure of several MPa \cite{azami2023development}.

A system with a single actuation source simplifies the control algorithm, is suitable for simple shapes made of high-tensile-strength materials, and downsizes the total system.
Numerous studies have employ a single actuation source, allowing an in-plane motion \cite{wei2022soft} or limiting out-of-plane motion through the use of a specific structure \cite{out-of-plane-scharff2019reducing,azami2023development} or optimization of design parameters \cite{out-of-plane-su2022optimizing,out-of-plane-li2023tailoring}.
The actuation force can be directly transformed into the deformation; however, flexibility and adaptability in the out-of-plane directions are sacrificed.

In this study, we developed a soft finger capable of adaptive-twist deformation and grasping-by-wrapping function using a single actuation source.
To achieve these two functions by a single actuation source, we propose a variable stiffness mechanism that can adaptively change the stiffness as the pressure becomes higher.
Although variable stiffness mechanisms have been proposed in the literature, including the jamming mechanism \cite{wall2015selective} and lock mechanism \cite{lock-structure-chung2019shape}, these studies mainly focused on variable stiffness in a single bending direction.
Other studies \cite{balljoint-sozer2023robotic, balljoint-onda2023tube} required multiple actuation sources.
Therefore, we propose a simple self-locking mechanism for the expansion of soft materials.
We determined a design parameter of the proposed mechanism based on a finite element analysis (FEA) result.
Basic experimental results and demonstrations of various objects being grasped are reported.

The rest of this paper is organized as follows.
\refsect{sect:finger} presents the structure of the proposed soft finger.
Then, \refsect{sect:variable_stiffness_evaluation} describes the design and fabrication of the proposed finger, including the parameter design based on the FEA and the validation of the FEA result by an experiment.
\refsect{sect:experiment} reports the experimental results obtained using the three fingered soft hand, demonstrating that various objects can be grasped.
We also conducted an experiment involving the picking task in a vegetable factory.
Finally, \refsect{sect:discussion} discusses the findings, and \refsect{sect:conclusion} summarizes the obtained results and concludes this paper.
\section{Soft Finger for Grasping by Wrapping\label{sect:finger}}

%%%%%%%%%%%%%%%%%%%%%%%%%%%%%%%%%%%%%%%%%%%%%%%%
\begin{figure}[tbp]
  \centering
    \includegraphics[width=\hsize]{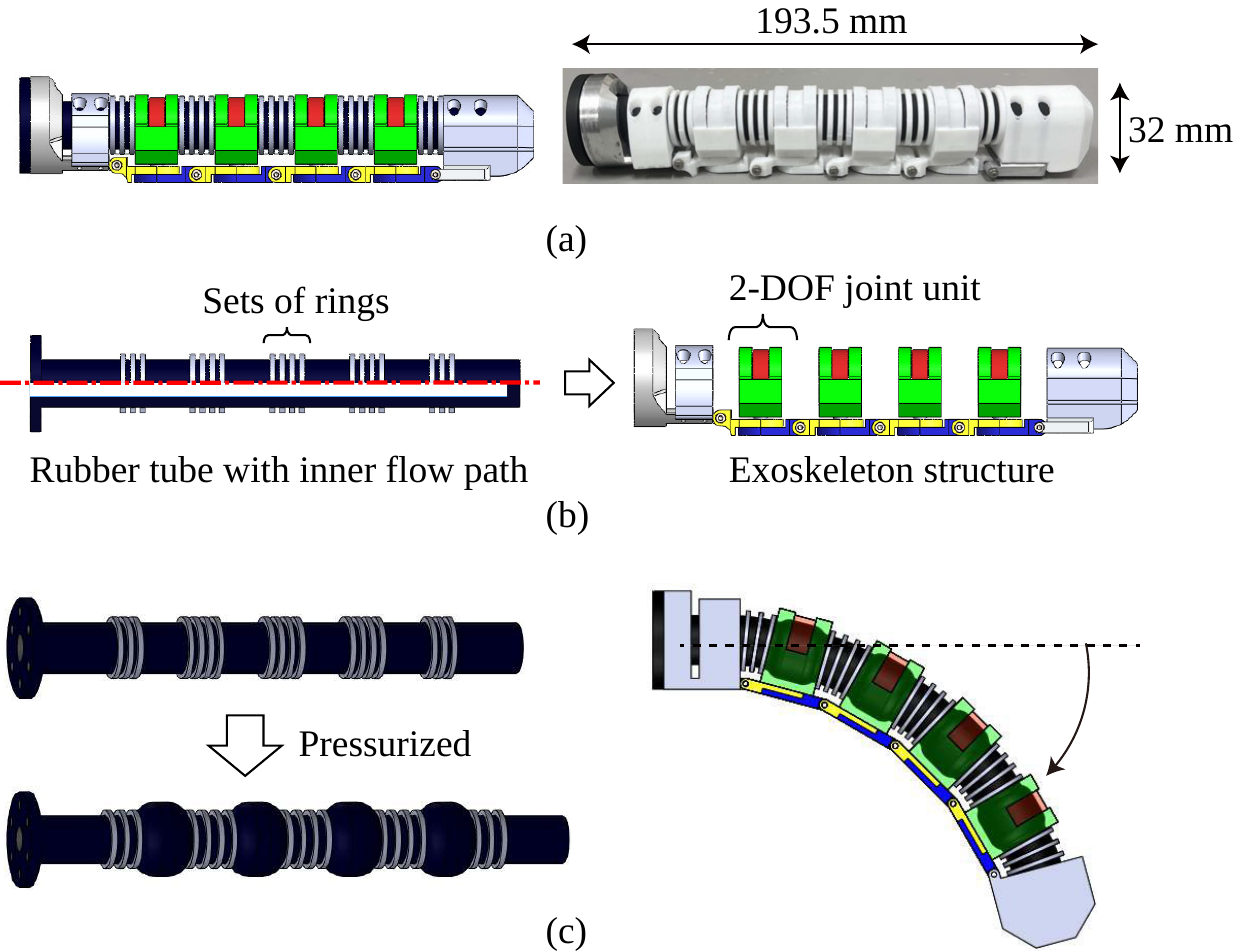}
  \caption{(a) Developed soft finger and (b) its structure.
  The finger consists of a rubber tube and an exoskeleton that converts internal pressure into desired finger movements, as shown in (c).}
  \label{fig:finger_structure}
\end{figure}
%%%%%%%%%%%%%%%%%%%%%%%%%%%%%%%%%%%%%%%%%%%%%%%%
\begin{figure}[t]
    \centering
    \includegraphics[width=0.8\hsize]{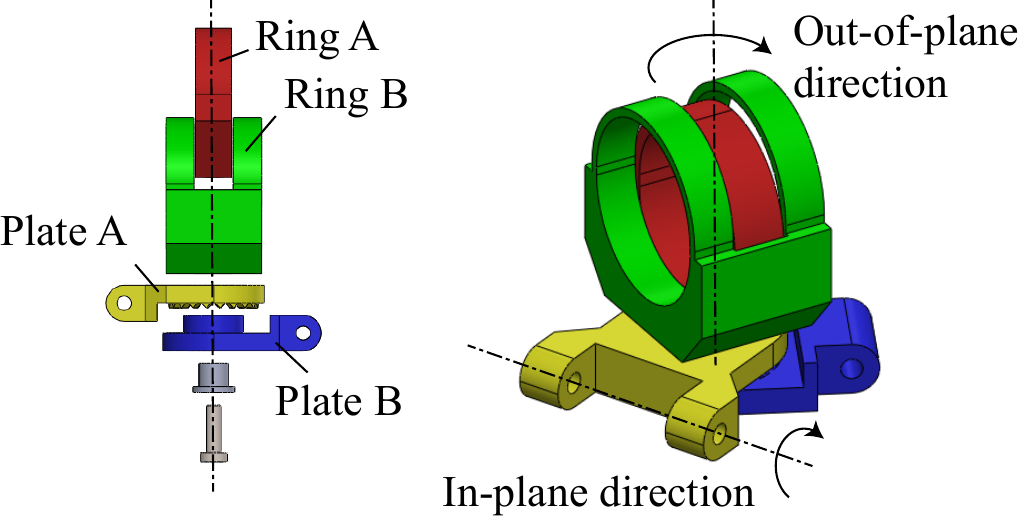}
  \caption{2-DOF joint unit in the exoskeleton, which enables adaptive-twist deformation in both in-plane and out-of-plane directions.}
  \label{fig:2dof_joint_unit}
\end{figure}
%%%%%%%%%%%%%%%%%%%%%%%%%%%%%%%%%%%%%%%%%%%%%%%%
%%%%%%%%%%%%%%%%%%%%%%%%%%%%%%%%%%%%%%%%%%%%%%%%
\begin{figure}[t]
    \centering
    \includegraphics[width=\hsize]{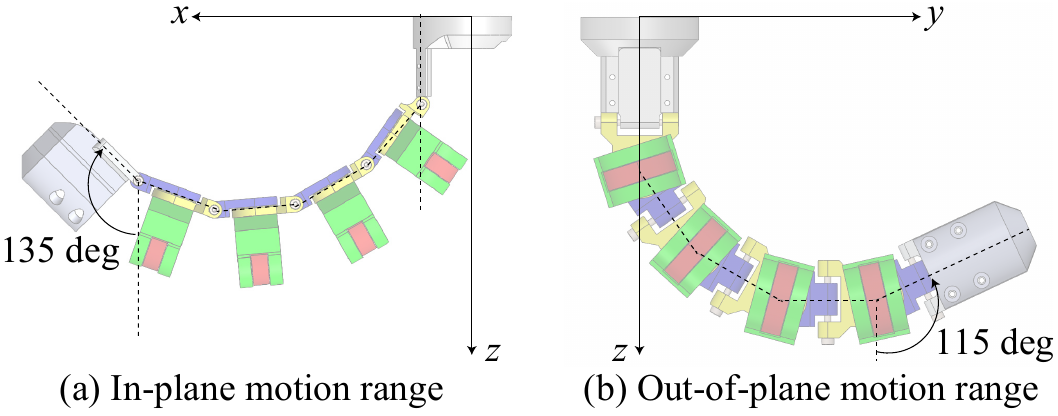}
  \caption{Motion range of the soft finger in (a) in-plane and (b) out-of-plane directions.}
  \label{fig:exoskeleton_structure}
\end{figure}
%%%%%%%%%%%%%%%%%%%%%%%%%%%%%%%%%%%%%%%%%%%%%%%%
\subsection{Soft Finger Structure for Adaptive-twist\label{subsect:structure_adaptive_twist}}
Figure \ref{fig:finger_structure}(a) shows the developed soft finger, which has a total length of 193.5 mm and a diameter of 32 mm.
When not pressurized, it exhibits flexibility in both in-plane and out-of-plane directions, as also shown in the accompanying video.
Figure \ref{fig:finger_structure}(b) illustrates an outline of its structure.
The finger consists of a rubber tube with inner flow path and an exoskeleton that converts the pressure inside the tube into desired finger movements.
The expansion of the tube in the radial direction is  partially constrained by rings.
The rings are arranged such that there is a space between each two sets of rings.
The expansion in this space is constrained by the green and red parts of the exoskeleton, as shown in \reffig{fig:finger_structure}(c).
These constraints allow the tube to elongate in the axial direction when its inside is pressurized, in a manner similar to that in \cite{suzumori1991flexible}.
As shown in \reffig{fig:finger_structure}(c), one side of the tube is constrained by an exoskeleton. 
Consequently, the finger bends toward the constrained side, which mainly results in the in-plane motion.

Figure \ref{fig:2dof_joint_unit} shows 2-DOF joint unit that constitutes the exoskeleton.
This unit has a variable stiffness function, which is explained in the next subsection.
Each unit has two plates, Plates A and B, as indicated by the yellow and blue parts in \reffig{fig:2dof_joint_unit}, respectively.
Plate A in one unit is connected to Plate B in the adjoining unit by a rotational joint along the $y$-axis, allowing the in-plane motion as shown in \reffig{fig:exoskeleton_structure}(a).
Moreover, Plates A and B in the same unit are connected by another rotational joint along the $x$-axis, which allows the out-of-plane motion as shown in \reffig{fig:exoskeleton_structure}(b).
Plates A and B exhibit a locking mechanism around the $x$-axis rotation.
This mechanism achieves variable stiffness.
The maximum bending angle with the NBR tube mounted in the 2-DOF joint units is approximately 135$^{\circ}$ in the in-plane direction, and 115$^{\circ}$ in the out-of-plane direction, as shown in \reffig{fig:exoskeleton_structure}.

\subsection{Variable Stiffness Mechanism for Grasping by Wrapping\label{subsect:structure_variable_stiffness}}
%%%%%%%%%%%%%%%%%%%%%%%%%%%%%%%%%%%%%%%%%%%%%%%%
\begin{figure}[t]
    \centering
    \subfloat[Interlock of protrusions between Plates A and B.\label{fig:variable_stiffness_mechanism_protrusions}]{
        \includegraphics[width=0.3\hsize]{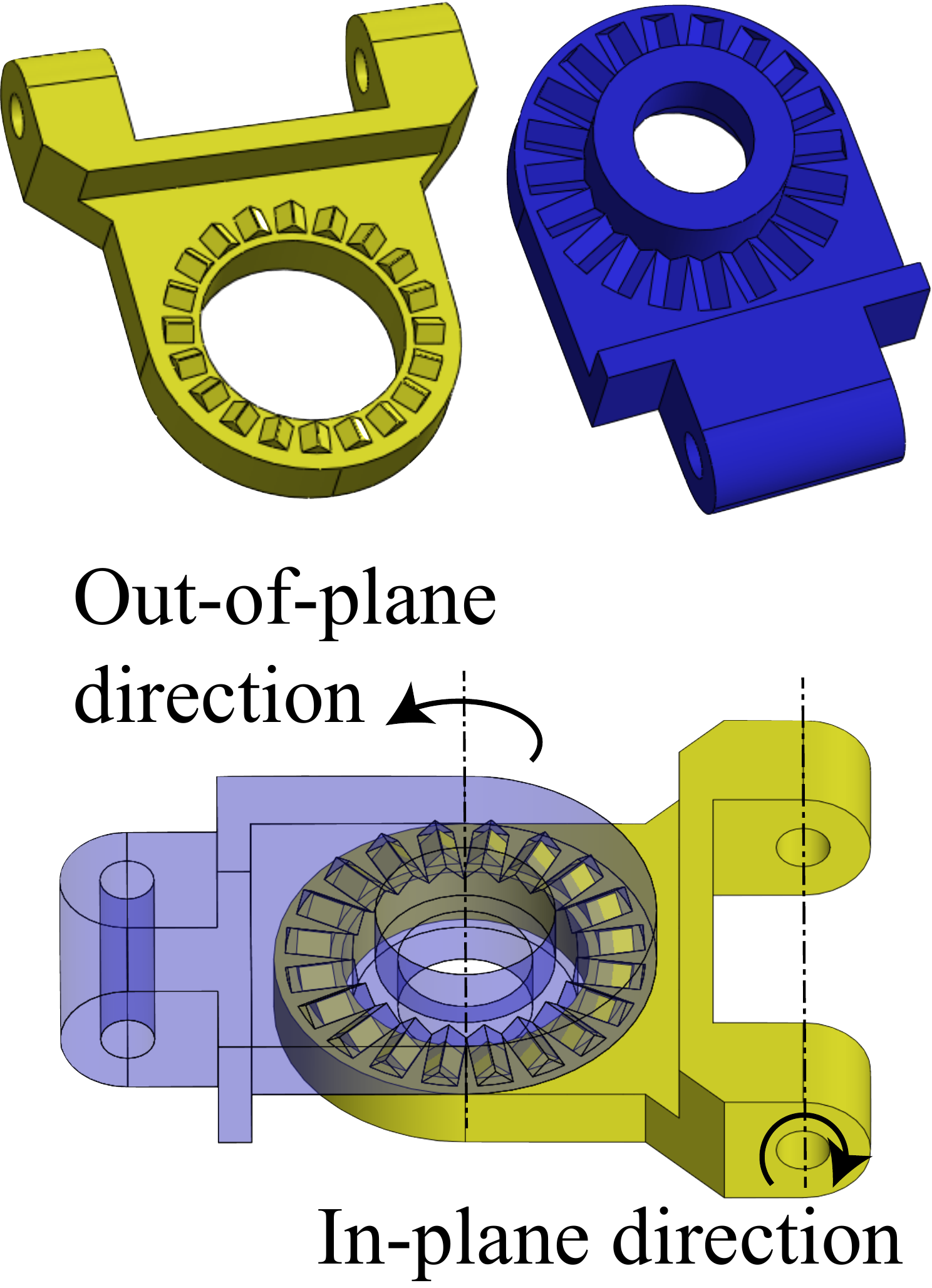}
    }
    \hspace{1em}
    \subfloat[Adaptive locking by inflation of the inner tube.\label{fig:variable_stiffness_mechanism_lock}]{
        \includegraphics[width=0.5\hsize]{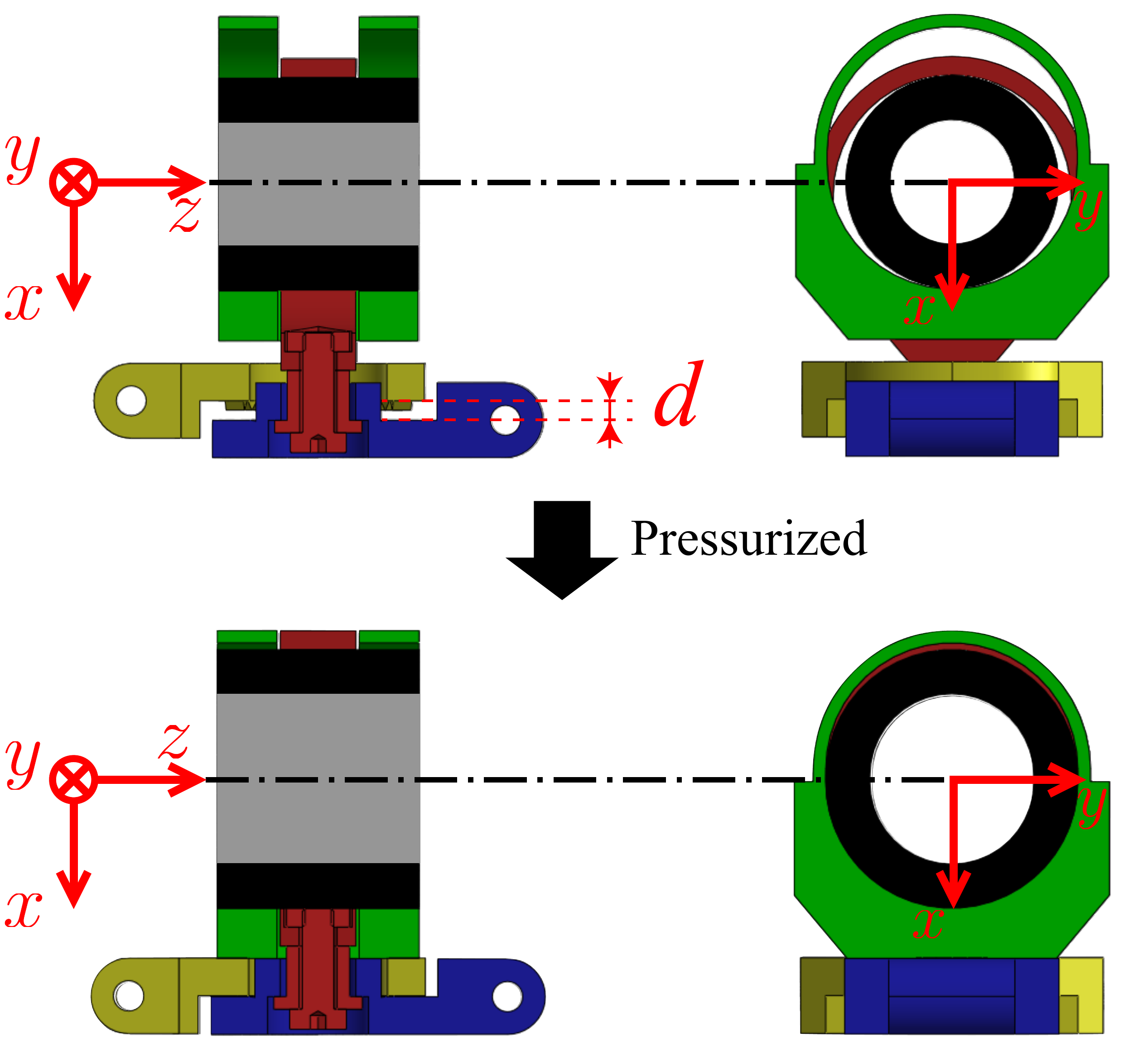}
    }
  \caption{Variable stiffness structure of the 2-DOF joint unit.}
  \label{fig:variable_stiffness_mechanism}
\end{figure}    
%%%%%%%%%%%%%%%%%%%%%%%%%%%%%%%%%%%%%%%%%%%%%%%%

Figure \ref{fig:variable_stiffness_mechanism} shows the details of the variable-stiffness mechanism. 
Ring A, indicated in red part, is directly connected to the rotational axis, whereas Ring B, indicated by the green part, is not.
When the tube is not pressurized, the holes in Rings A and B are not concentric, and Ring B does not contact Plate A.
When the tube is pressurized, the inflation of the tube pushes these two rings from inside such that the centers of the ring holes align. 
Consequently, Ring B is pushed downward into contact with Plate A.
At the same time, Plate A is also pushed downward.
Eventually, the locking mechanism between Plates A and B is activated, resulting in high rigidity in the $x$-axis rotation.
This mechanism achieves sufficient rigidity in directions tangential to the contact surface when grasping an object, as illustrated in \reffig{fig:wrapping_grasp_concept_tangent_direction}.

%%%%%%%%%%%%%%%%%%%%%%%%%%%%%%%%%%%%%%%%%%%%%%%%
\begin{table}[t]
    \centering
    \caption{Comparison with variable stiffness mechanisms in soft robots.}
    \label{tab:comparison_with_related_works}
    \begin{tabular}{c|l}
        Methodology & Difference from this study \\ \hline
         Jamming \cite{wei2022soft,Crowley2022RAL} & Limited in the in-plane direction \\
         Locking \cite{lock-structure-chung2019shape} or switching \cite{Bastien2023TRO} & Limited in the in-plane direction \\
         Inner-chamber inflation \cite{balljoint-sozer2023robotic, balljoint-onda2023tube} & Requires additional actuation \\
    \end{tabular}
\end{table}
%%%%%%%%%%%%%%%%%%%%%%%%%%%%%%%%%%%%%%%%%%%%%%%%
Table \ref{tab:comparison_with_related_works} compares the proposed mechanism with those presented in related works.
References \cite{wei2022soft,Crowley2022RAL} achieved variable stiffness using jamming mechanism, but the deformation of those soft hands was limited in the in-plane direction.
Other studies utilizing locking \cite{lock-structure-chung2019shape} or switching mechanism \cite{Bastien2023TRO} were also limited in the in-plane deformation.
In particular, \cite{Bastien2023TRO} utilized a reconfigurable kinematic structure but required an additional actuation for switching the configuration.
Variable stiffness mechanisms based on inner-chamber inflation also required separate pressure control systems \cite{balljoint-sozer2023robotic, balljoint-onda2023tube}.
Our mechanism does not require an additional actuation source for the variable stiffness.
The stiffness adaptively changes as a single actuation source generates sufficient pressure for in-plane motion, allowing us to use a simple control algorithm.
\section{Design of Variable Stiffness Mechanism\label{sect:variable_stiffness_evaluation}}
\subsection{Fabrication of Soft Finger}
Most parts of the exoskeleton are fabricated using a 3D printer with polylactic acid (PLA).
Some parts that required high strength, including the finger root and tip, were made of A7075.
In this paper, we used oil as the driving fluid for higher output and selected NBR (Shore A hardness 70) as the tube material because of its oil resistance and high tensile strength. 

In the 2-DOF joint unit, the distance between Plates A and B, denoted by $d$ in \reffig{fig:variable_stiffness_mechanism}, is an important design parameter that determines the variable stiffness profile.
In the following subsections, we determine $d$ considering the grasping of an object by wrapping and conducting an FEA.

\subsection{Design Parameter of Variable Stiffness Mechanism}
\subsubsection{Moment Required in 2-DOF Joint Unit for Grasping by Wrapping\label{eq:moment_required}}
%%%%%%%%%%%%%%%%%%%%%%%%%%%%%%%%%%%%%%%%%%%%%%%%
\begin{figure}[t]
    \centering
    \subfloat[\label{fig:required_function_wrapping_image}]{
        \includegraphics[width=0.35\hsize]{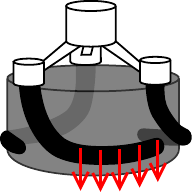}
    }
    \subfloat[\label{fig:required_function_force_approximation}]{
        \includegraphics[width=0.45\hsize]{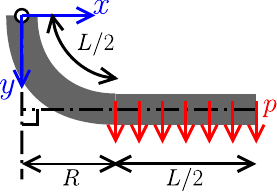}
    }
  \caption{Moment required in the 2-DOF joint unit for grasping by wrapping.}
  \label{fig:required_function}
\end{figure}
%%%%%%%%%%%%%%%%%%%%%%%%%%%%%%%%%%%%%%%%%%%%%%%%
We estimated the moment required in the 2-DOF joint unit for grasping an object by wrapping.
For simplicity, we considered the scenario shown in \reffig{fig:required_function_wrapping_image}, where fingers grasp a cylindrical object by wrapping it.
Then, we calculated the moment required in the 2-DOF joint unit placed in the root of each finger, which required the largest moment.
Let us suppose that the half of the finger bends by 90$^\circ$ in the out-of-plane direction, and the other half wraps the object, as shown in \reffig{fig:required_function_force_approximation}.

Further, let $L$ denote the length of the finger, and $R$ denote the radius of the bending part, which satisfies $R = L / \pi$ .
We assumed that the grasping force was uniformly distributed in the contact area.
The grasping force per unit length, denoted by $p$, was calculated as
\begin{align}
  p = \frac{2mg}{nL}
\end{align}
where $m$ is the mass of the object, $g$ is the gravitational acceleration, and $n$ is the number of fingers.
Then, the moment applied to the root of the fingers, denoted by $M$, can be calculated as
\begin{align}
 M = \int_{R}^{R + \frac{L}{2}}px\,dx = \frac{mgL}{n}\left(\frac{1}{\pi} + \frac{1}{4}\right) .
 \label{eq:variable_stiffness_required_moment_integral}
\end{align}

Assuming a practical application presented in Section \ref{sect:experiment_cabbages}, we set $m = 1.5 \,\mathrm{kg}$ as the target value.
Also, we set $L \simeq 200 \,\mathrm{mm}$ referring to \cite{azami2023development}.
Setting $n=3$ as the minimum number for a stable grasping, we can calculate the moment as
\begin{align}
  M \simeq M_{\mathrm{required}} := 0.6 \,\mathrm{Nm} . \label{eq:variable_stiffness_required_momea}
\end{align}

\subsubsection{FEA\label{subsect:FEM_evaluation}}
%%%%%%%%%%%%%%%%%%%%%%%%%%%%%%%%%%%%%%%%%%%%%%%%
\begin{figure}[t]
  \centering
  \includegraphics[width=1.0\hsize]{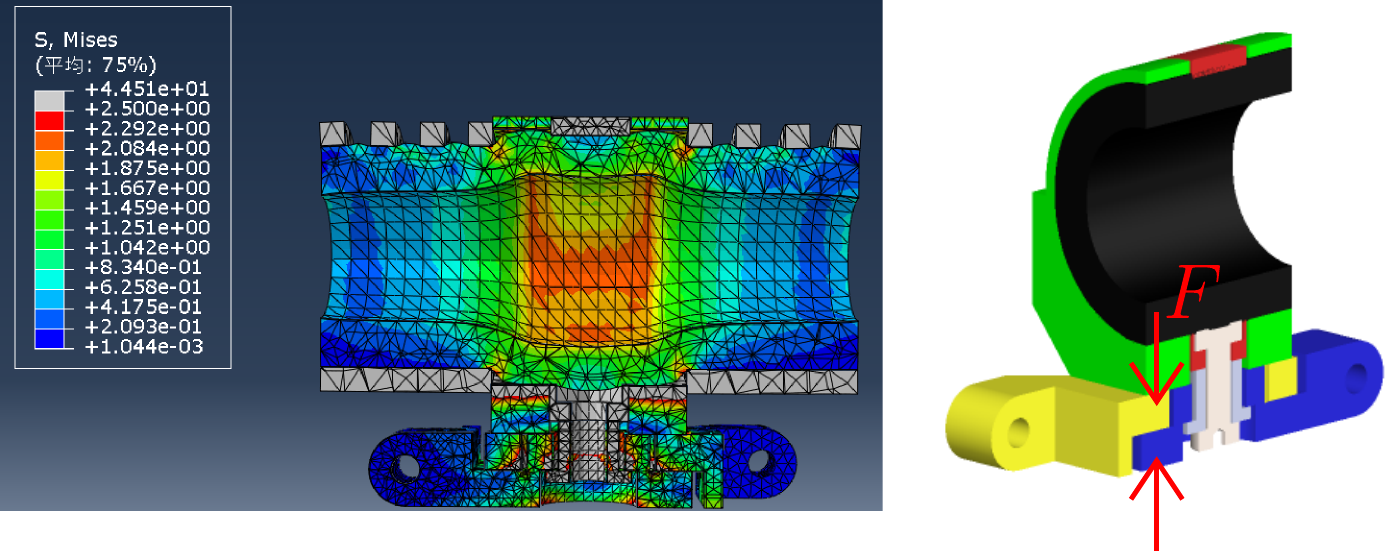}
  \caption{FEA of the variable stiffness mechanism to calculate the force $F$ between Plates A and B.}
  \label{fig:FEM_illustration}
\end{figure}
%%%%%%%%%%%%%%%%%%%%%%%%%%%%%%%%%%%%%%%%%%%%%%%%
%%%%%%%%%%%%%%%%%%%%%%%%%%%%%%%%%%%%%%%%%%%%%%%%
\begin{figure}
  \centering
    \includegraphics[width=\hsize]{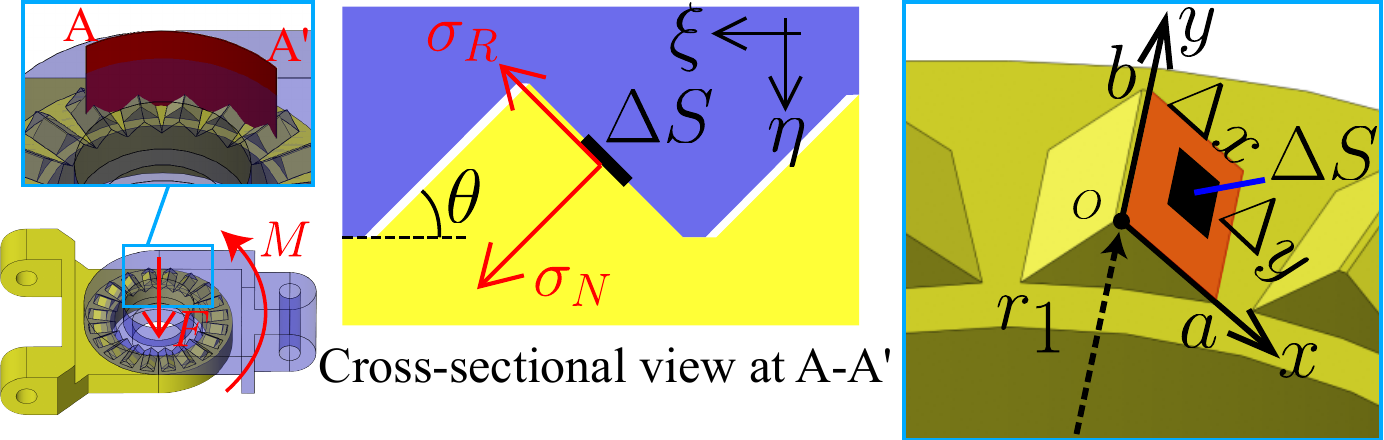}
  \caption{Calculation of moment $M$ from the force $F$ between Plates A and B. We calculate $M$ considering the force at the contact surface $S$ on a single protrusion.}
  \label{fig:hinge_protrusion}
\end{figure}
%%%%%%%%%%%%%%%%%%%%%%%%%%%%%%%%%%%%%%%%%%%%%%%%

We calculated the maximum moment $M_{\max}$, where the variable stiffness mechanism can represent different values of the design parameter $d$.
First, force $F$ between Plates A and B when the rubber tube is pressurized is estimated using FEA, as shown in \reffig{fig:FEM_illustration}. 
FEA was performed using Abaqus, in which
the Mooney-Rivlin model was employed to model the behavior of the NBR material \cite{gent2012engineering}.
In the figure, the color indicates the value of Mises stress.
The maximum values of the stress in the NBR and PLA parts are 2.5 MPa and 39 MPa, respectively.
These values are below the tensile strength of the materials: 12.7 MPa for NBR and 55 MPa for PLA.
Subsequently, $M_{\max}$ was calculated from the estimated value of $F$ as follows.

As shown in \reffig{fig:hinge_protrusion}, we consider applying an external moment $M$ whereas Plates A and B are interlocked with a pressing force $F$. Focusing on the top surface $S$ of one of the protrusions of Plate A, highlighted in red in \reffig{fig:hinge_protrusion}, we define an $xy$ coordinate system on surface $S$. We then consider a rectangular region $\Delta S$ centered at position $(x,y)$ with a width $\Delta x$ and height $\Delta y$. If $\sigma_{N}$ and $\sigma_{R}$ denote the static friction force and normal force per unit area on $\Delta S$, respectively, the net static friction force and normal force on $S$, denoted by $N$ and $R$, respectively, can be calculated as follows:
\begin{align}
  N = \int_{0}^{a}\int_{0}^{b}\sigma_{N}\,dydx
  \label{eq:N_integral},
  \quad
  R = \int_{0}^{a}\int_{0}^{b}\sigma_{R}\,dydx
  %\label{eq:R_integral}
\end{align}
where $a$ and $b$ represent the width and height of $S$, respectively.

Next, we transform $\sigma_N$ and $\sigma_R$ to those in newly defined $\xi$ and $\eta$ axes as shown in the cross-sectional view in \reffig{fig:hinge_protrusion}, which are denoted by $\sigma_\xi$ and $\sigma_\eta$.
The relationship between $(\sigma_N, \sigma_R)$ and $(\sigma_\xi, \sigma_\eta)$ is given as
\begin{align}
  \sigma_{N} = \sigma_{\xi}\sin\theta + \sigma_{\eta}\cos\theta
  %\label{eq:N_eta_xi_component},
  \quad
  \sigma_{R} = \sigma_{\xi}\cos\theta - \sigma_{\eta}\sin\theta
  \label{eq:R_eta_xi_component}
\end{align}

From the equilibrium of the forces acting on Plate A and the equilibrium of the moments around the joint axis, we obtain
\begin{align}
  F = n\int_{0}^{a}\int_{0}^{b}\sigma_\eta \,dydx\label{eq:F_integral}
  \\
  M = n\int_{0}^{a}\int_{0}^{b}d(x,y) \cdot \sigma_\xi \,dydx\label{eq:M_integral}
\end{align}
where $n$ is the total number of the protrusions. 
$d(x,y)$ is the distance between a point $(x,y)$ on surface $S$ and the rotational axis, which is given by:
\begin{align}
  d(x,y) = \sqrt{(x \cos\theta)^2 + (y + r_1)^2}
\end{align}
where $\theta$ is the inclination angle of the protrusions, and $r_1$ is the distance between the joint axis and the origin $O$ of the $xy$ coordinate system as shown in \reffig{fig:hinge_protrusion}.

The condition for preventing the two plates from slipping is given as:
\begin{align}
\textcolor{\eq_color}{
  R \leq \mu N\label{eq:no_slip}
}
\end{align}
where $\mu$ is the static friction coefficient between Plates A and B.
Substituting \refeq{eq:N_integral} -- \refeq{eq:R_eta_xi_component} into \refeq{eq:no_slip}, we obtain
\begin{align}
  \int_{0}^{a}\int_{0}^{b}\sigma_{\xi}\,dydx
  \leq \frac{\sin\theta + \mu\cos\theta}{\cos\theta - \mu\sin\theta}
  \int_{0}^{a}\int_{0}^{b}\sigma_{\eta}\,dydx\label{eq:no_slip_integral}
\end{align}
Note that if $\cos\theta - \mu\sin\theta \leq 0$, interlocking between Plates A and B will be maintained regardless of force $F$. This is inappropriate for the variable stiffness mechanism's function because it means that there is no flexibility in the out-of-plane direction. Therefore, we set the design requirement for $\mu$ and $\theta$ such that $\cos\theta - \mu\sin\theta > 0$.

Assuming that $\sigma_{R}$ and $\sigma_{N}$ are constant in $S$, we can rewrite \refeq{eq:M_integral} as
\begin{align}
  \sigma_{\xi} = \frac{M}{nD},
    \quad
    D: = \displaystyle{\int_{0}^{a}\int_{0}^{b}d(x,y)\,dydx}
  \label{eq:sigma_xi_approx}
\end{align}
Substituting \refeq{eq:F_integral} and \refeq{eq:sigma_xi_approx} into \refeq{eq:no_slip_integral}, we obtain
\begin{align}
\textcolor{\eq_color}{
  M \leq M_{\max} := F\cdot\frac{\sin\theta + \mu\cos\theta}{\cos\theta - \mu\sin\theta}\cdot\frac{D}{ab}
  }\label{eq:max_moment}
\end{align}
As long as \refeq{eq:max_moment} holds, the $x$-axis rotational joint of the finger remains locked, increasing rigidity of the finger in the out-of-plane direction.

We set the range of the static friction coefficient as $0.4 \le \mu \le 0.7$ based on \cite{pla_friction-zhang2020friction}, considering the 100\% infill density for 3D printing.
We calculated $M_{\max}$ using these values and the value of $F$ obtained from the FEA result.

%%%%%%%%%%%%%%%%%%%%%%%%%%%%%%%%%%%%%%%%%%%%%%%%
\begin{figure}[t]
  \centering
  \includegraphics[width=0.9\hsize]{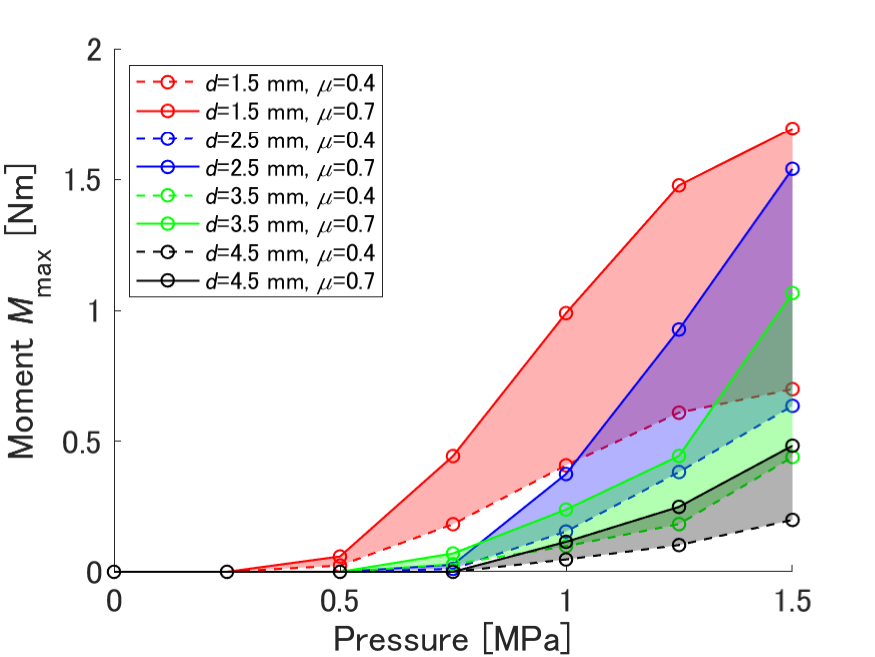}
  % \vspace*{-4mm}
  \caption{Maximum moments $M_{\max}$ that the variable stiffness unit can withstand for various values of $d$.}
  \label{fig:fem_pressure_moment}
\end{figure}
%%%%%%%%%%%%%%%%%%%%%%%%%%%%%%%%%%%%%%%%%%%%%%%%

\subsubsection{Design of Parameter $d$}
\reffig{fig:fem_pressure_moment} shows the results of $M_{\max}$ calculated for four different values of the design parameter: $d = 1.5, 2.5, 3.5, 4.5$ [mm].
It was observed that $M_{\max}$ increased as the pressure increased, generating the function of variable stiffness.
Also, the parameter $d$ has a large effect on the profile of the variable stiffness.

We assumed a nominal pressure value of 1.5 MPa.
Therefore, we determined the profiles for the variable stiffness as follows: 
\begin{itemize}
  \item For lower pressure ($P \leq 0.75 \mathrm{MPa}$), $M_{\max}$ should be sufficiently small to ensure that the finger is soft and adaptive to the environment.
  \item At higher pressures ($P \simeq 1.5 \mathrm{MPa}$), $M_{\max} > M_{\mathrm{required}} = 0.6 \,\mathrm{Nm}$ is required for grasping.
\end{itemize}
To satisfy the aforementioned profile, we selected $d = 2.5$ [mm] in an actual 2-DOF joint unit.

\subsection{Experimental Validation\label{subsect:experimental_validation}}
%%%%%%%%%%%%%%%%%%%%%%%%%%%%%%%%%%%%%%%%%%%%%%%%
\begin{figure}[t]
  \centering
  \includegraphics[width=\hsize]{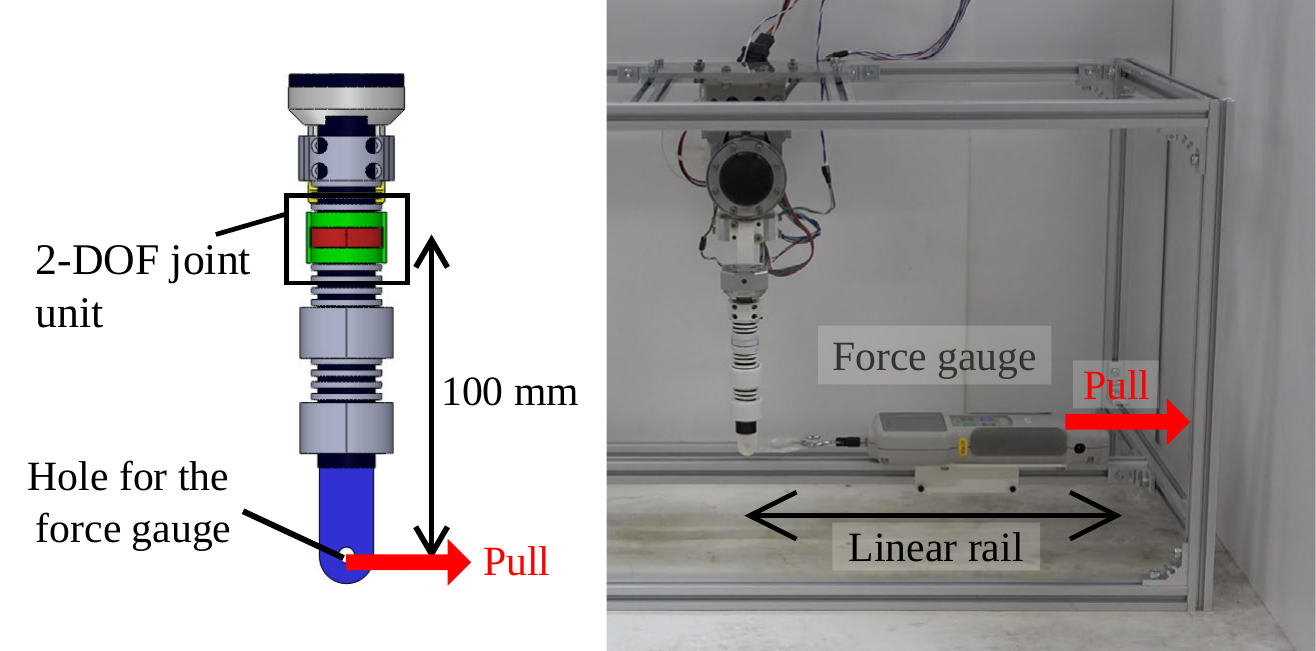}
  \caption{Experimental setup to evaluate the variable stiffness mechanism.}
  \label{fig:variable_stiffness_evaluation_experimental_setup}
\end{figure}
%%%%%%%%%%%%%%%%%%%%%%%%%%%%%%%%%%%%%%%%%%%%%%%%
%%%%%%%%%%%%%%%%%%%%%%%%%%%%%%%%%%%%%%%%%%%%%%%%
\begin{figure}[t]
  \centering
  \includegraphics[width=0.8\hsize]{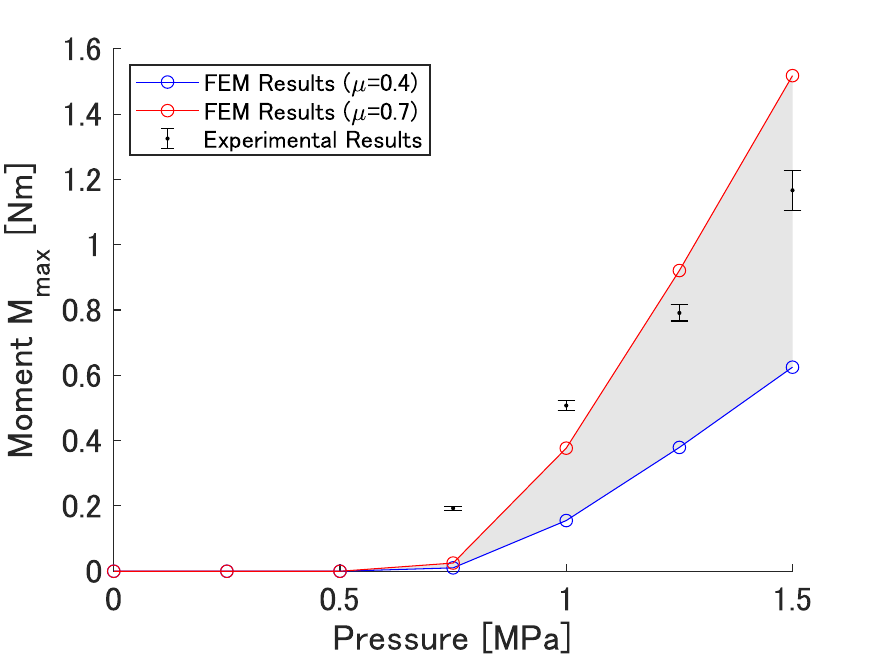}
  \caption{Maximum moment $M_{\max}$ that the variable stiffness mechanism can withstand for different pressure values. The gray area is the estimated result in \reffig{fig:fem_pressure_moment} with $d$=2.5 mm. 
  }
  \label{fig:pressure_moment_overwrap_graph}
\end{figure}
%%%%%%%%%%%%%%%%%%%%%%%%%%%%%%%%%%%%%%%%%%%%%%%%
We conducted an experiment to validate the maximum moment values estimated in the previous subsection.
\reffig{fig:variable_stiffness_evaluation_experimental_setup} shows the experimental setup using a finger consisting of a single variable stiffness mechanism. 
The fingertip was pulled using a force gauge (IMADA's ZP-50N). 
The applied moment was calculated by multiplying the measured force by the distance between the tip and the rotational center of the variable stiffness mechanism, which was 100 mm.
As shown in \reffig{fig:variable_stiffness_evaluation_experimental_setup}, the force gauge was set on an aluminum frame such that it moved only in a linear direction, and pulled to apply lateral force to the finger.
The measurements were conducted at five different pressure values between 0.5 and 1.5 MPa with 0.25 MPa interval. 
We used a hydraulic pump \cite{komagata2019design} to pressurize the finger and recorded the maximum force observed before the lock of the variable stiffness mechanism was unlocked by the external force.
This procedure was repeated 10 times for each pressure value.

%\subsubsection{Result}
The experimental results are shown in \reffig{fig:pressure_moment_overwrap_graph}, overlaid with the result of $M_{\max}$ estimated for $d=2.5$ [mm] in \reffig{fig:fem_pressure_moment}.
In the figure, each dot indicates the average of 10 trials at each pressure, and the error bar indicates the standard error.
The corresponding plot when the pressure was lower than 0.5 MPa can be regarded as 0 Nm in the figure because Plates A and B were not in contact, and the locking mechanism was not activated. 

In \reffig{fig:pressure_moment_overwrap_graph}, the experimental results are inside the gray area when $P \ge$ 1.25 MPa, which is the range typically used for grasping an object. 
However, the experimental results were outside the gray area when $P \le$ 1 MPa.
The reason is that Plates A and B were assumed to be parallel in the FEA, whereas these plates were not parallel but were in contact with each other for a smaller force $F$ with lower pressure.
This contact resulted in stronger locking between the two plates than in the simulation.

From the experiment, we found that the maximum moment $M_{\max}$ was approximately 1.2 Nm for 1.5 MPa drive pressure.
This corresponds to 8.45 N (= 1.2 Nm / 142 mm) force acting on the finger tip because the distance between the first joint unit and finger tip is 142 mm in the actual finger.

\section{Object Grasping Experiment\label{sect:experiment}}
\subsection{Experimental Setup\label{subsect:robotarm_setup}}
%%%%%%%%%%%%%%%%%%%%%%%%%%%%%%%%%%%%%%%%%%%%%%%%
\begin{figure}[t]
  \centering
  \includegraphics[width=\hsize]{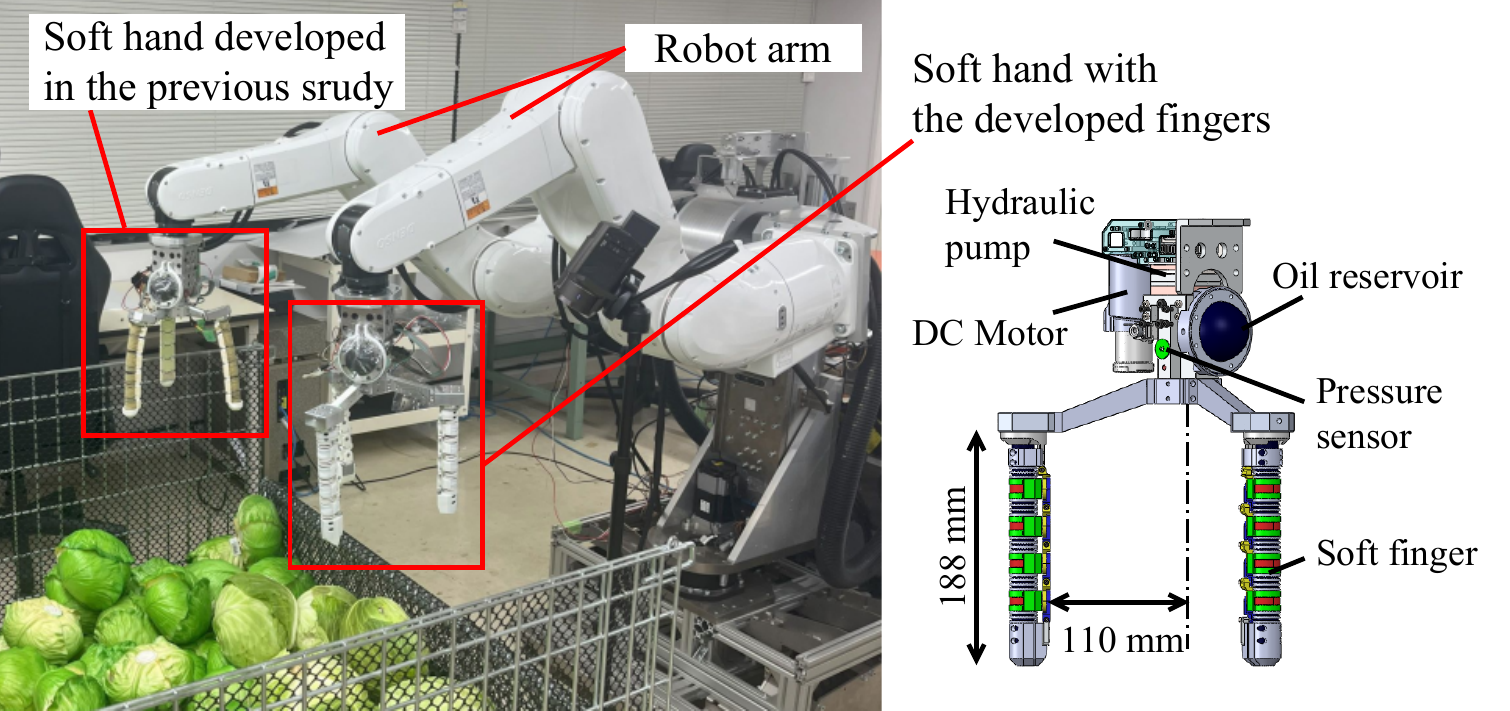}
  \caption{Dual-arm robot equipped with the soft hand consisting of the three fingers developed in this study. For a comparative experiment, one arm was equipped with a hand with the previously-developed fingers \cite{azami2023development}.}
  \label{fig:robot_arm_setup}
\end{figure}
%%%%%%%%%%%%%%%%%%%%%%%%%%%%%%%%%%%%%%%%%%%%%%%%
We conducted experiments to validate the grasping-by-wrapping function of the developed finger.
The scenarios for the following experiments are also shown in the accompanying video.
We developed a hand that consists of three fingers, as shown in the right panel of \reffig{fig:robot_arm_setup}.
The configuration of the hand is similar to the one developed in a previous study \cite{azami2023development}, consisting of a built-in hydraulic pump \cite{komagata2019design}, an oil reservoir and the developed soft fingers.
The total weight of the hand is approximately 2.6 kg, including the actuation unit, which consists of the pump, electric motor and oil reservoir.
The weight of the actuation part is 1.67 kg, which is 64\% of the total weight.
Note that a single pump actuates the three fingers.

As shown in \reffig{fig:robot_arm_setup}, the soft hand was mounted on a dual-arm robot, in which each arm was a DENSO WAVE VS087.
Using this system, we performed experiments that included the task of grasping a cabbage from multiple cabbages densely contained in a box.
For a comparison, we mounted the hand previously developed in \cite{azami2023development} onto the other arm.

\subsection{Basic Validation of Twisting and Wrapping Motion
\label{subsect:2ways_grabbing}}
%%%%%%%%%%%%%%%%%%%%%%%%%%%%%%%%%%%%%%%%%%%%%%%%
\begin{figure}[t]
  \centering
    \subfloat[\label{fig:no_object_c}]{
        \includegraphics[width=0.3\hsize]{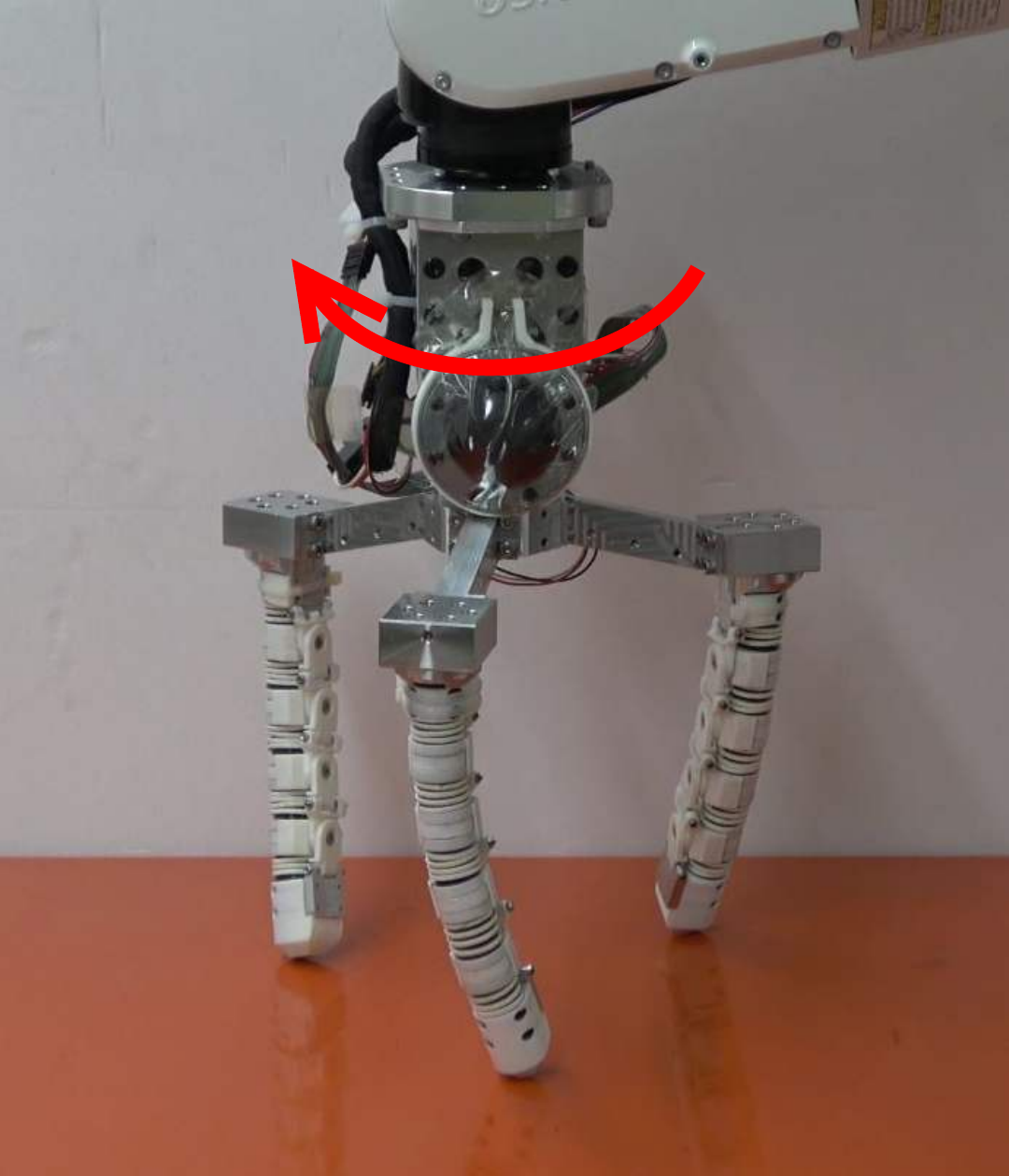}
    }
    \subfloat[\label{fig:no_object_e}]{
        \includegraphics[width=0.3\hsize]{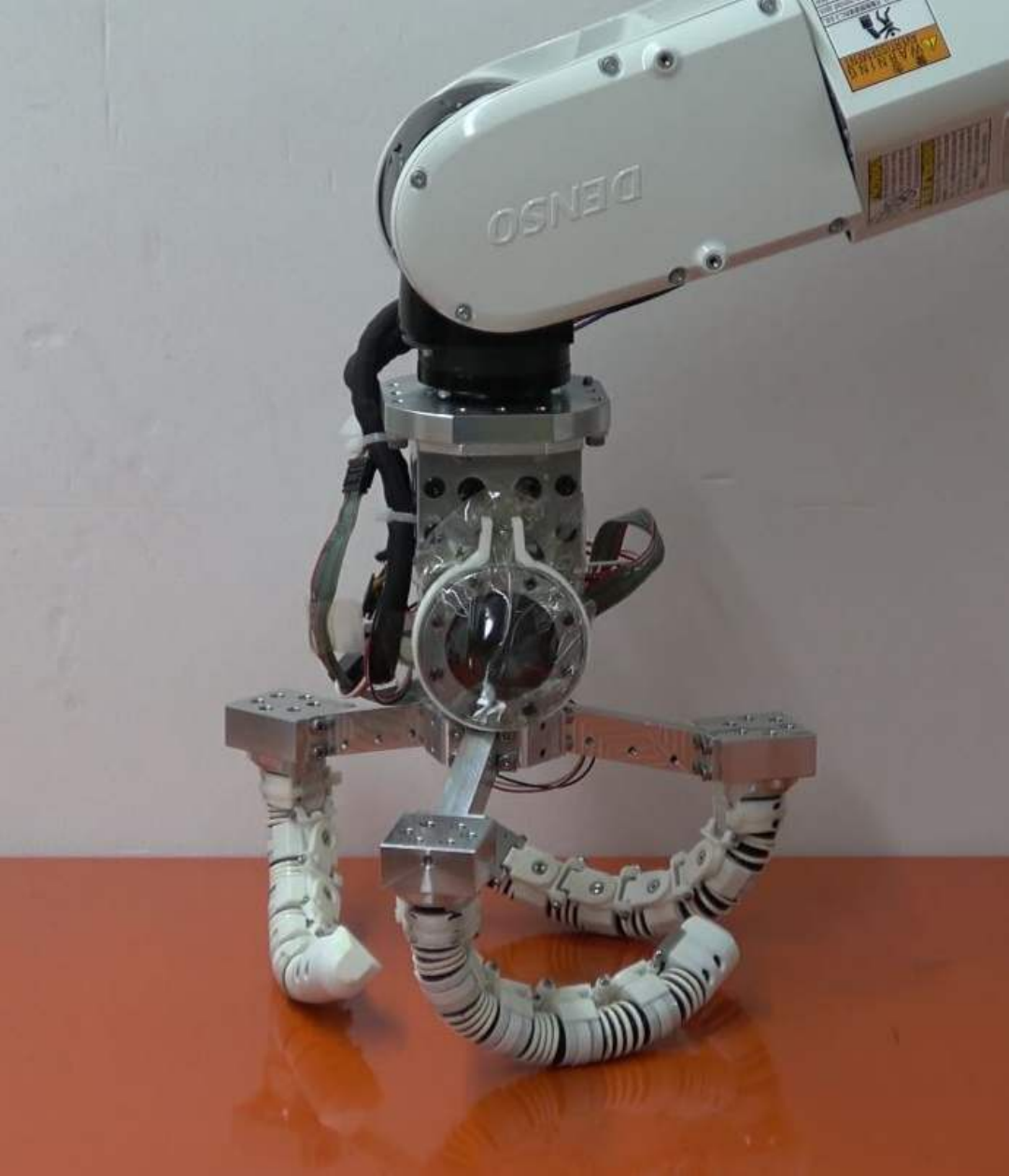}
    }
    \subfloat[\label{fig:no_object_f}]{
        \includegraphics[width=0.3\hsize]{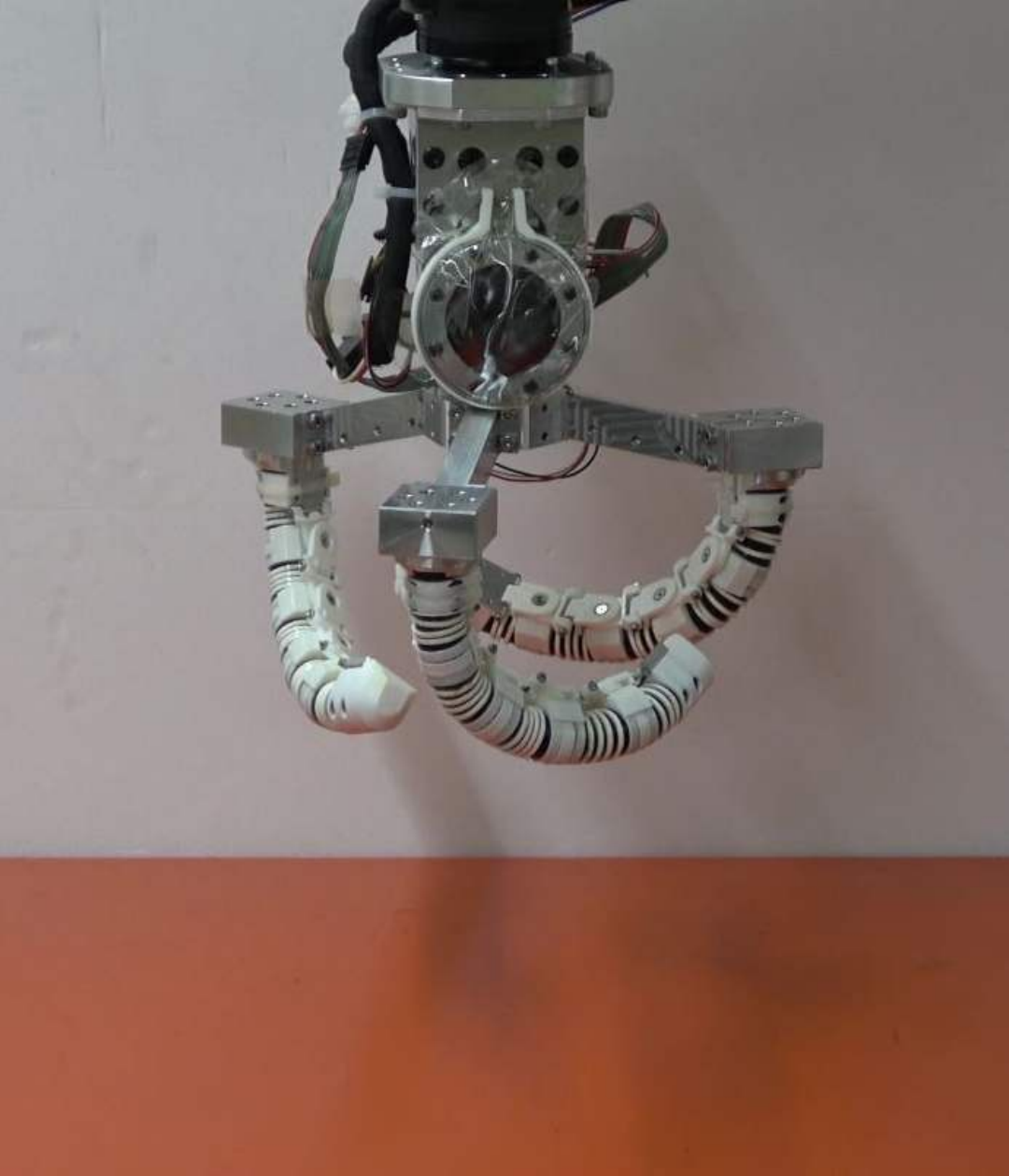}
    }
  \caption{Basic demonstration of twisting and wrapping motion. Each finger adaptively deforms after the finger tips contact. After pressurizing, a deformation for wrapping an object is activated as the variable stiffness mechanism.}
  \label{fig:no_object}
\end{figure}
%%%%%%%%%%%%%%%%%%%%%%%%%%%%%%%%%%%%%%%%%%%%%%%%
First, we demonstrated the adaptive-twisting and wrapping motions of the proposed mechanism.
Figure \ref{fig:no_object} demonstrates the use of a soft hand and a robot arm.
After the fingertips contacted, the wrist joint was rotated around the yaw axis, as indicated by the red arrow in \reffig{fig:no_object_c}.
Subsequently, the soft hand was moved to a lower position again.
As shown in \reffig{fig:no_object_e}, the soft fingers adaptively bent in the out-of-plane direction.
Finally, the soft fingers were pressurized, as shown in \reffig{fig:no_object_f}, which activated the variable stiffness mechanism. As a result, the soft fingers maintained the out-of-plane bending angle, while also bending in the in-plane direction even when the hand was moved back to a higher position. 
This process allowed the finger to wrap around a cylindrical object, as shown in the next subsection.

%%%%%%%%%%%%%%%%%%%%%%%%%%%%%%%%%%%%%%%%%%%%%%%%
\begin{figure}[t]
  \centering
        \includegraphics[width=\hsize]{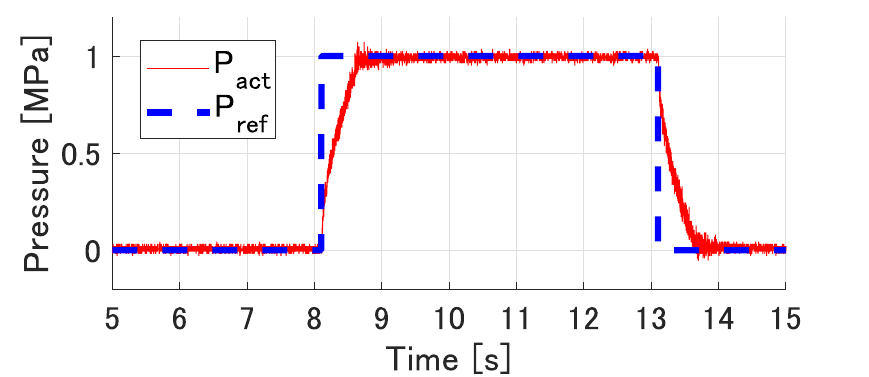}
    \caption{Step response of the pressure control with 1 MPa reference, starting from the state shown in \reffig{fig:no_object}(b) with twisted deformation. 
    \label{fig:step_response}}
\end{figure}
%%%%%%%%%%%%%%%%%%%%%%%%%%%%%%%%%%%%%%%%%%%%%%%%
We also measured the step response of the pressure control with a 1 MPa reference value.
Figure \ref{fig:step_response} shows the results of the step response, starting from the state shown in \reffig{fig:no_object_e}, with twisted deformation.
Assuming a first-order lag system, the time constant was calculated as 0.3 s.
A similar response was also obtained when the finger was bent only in the in-plane direction.
This result suggests that the proposed system has sufficient response speed regardless of the twisted deformation.

The response time of hydraulic actuation is generally slower than that of pneumatic actuation.
In particular, the hydraulic actuation system in this study does not incorporate any valve control whereas most pneumatic actuators achieve higher response time (rise time) using valve control.
From Fig. \ref{fig:step_response}, the rise time is approximately 0.6 s.
We consider this is comparable with pneumatic actuators because 0.5 s -- 1 s rise time is usually considered for such a high pressure \cite{Xavier2021}.
Under 1 MPa pressure control, it took approximately 3 s to close the fingers, as demonstrated in the accompanying video.
We consider this is also comparable with pneumatic actuators because the rise time in the angle control using a pneumatic actuator is typically 2.5 s -- 7.5 s \cite{Hyatt2019}.

\subsection{Grasping Objects by Wrapping\label{sect:various_objects_grabbing}}
\begin{figure}[t]
  \centering
    \subfloat[\label{fig:grab_various_objects_cylinder}]{
        \includegraphics[width=0.4\hsize]{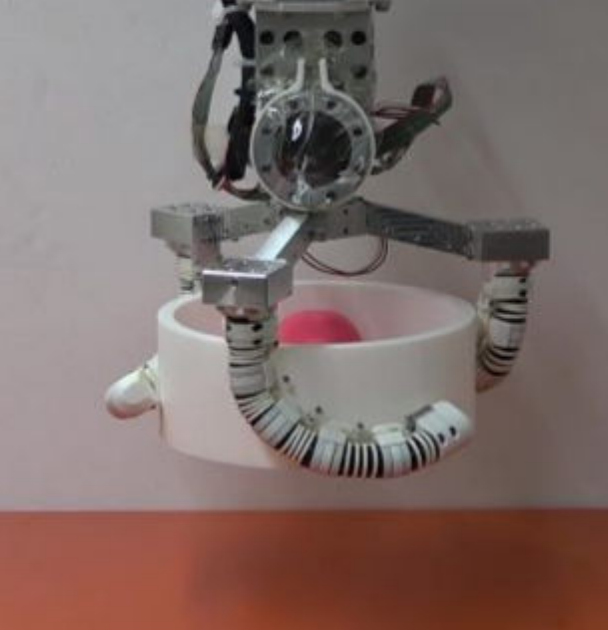}
    }
    \subfloat[\label{fig:1DOF_constraint_cylinder}]{
        \includegraphics[width=0.4\hsize]{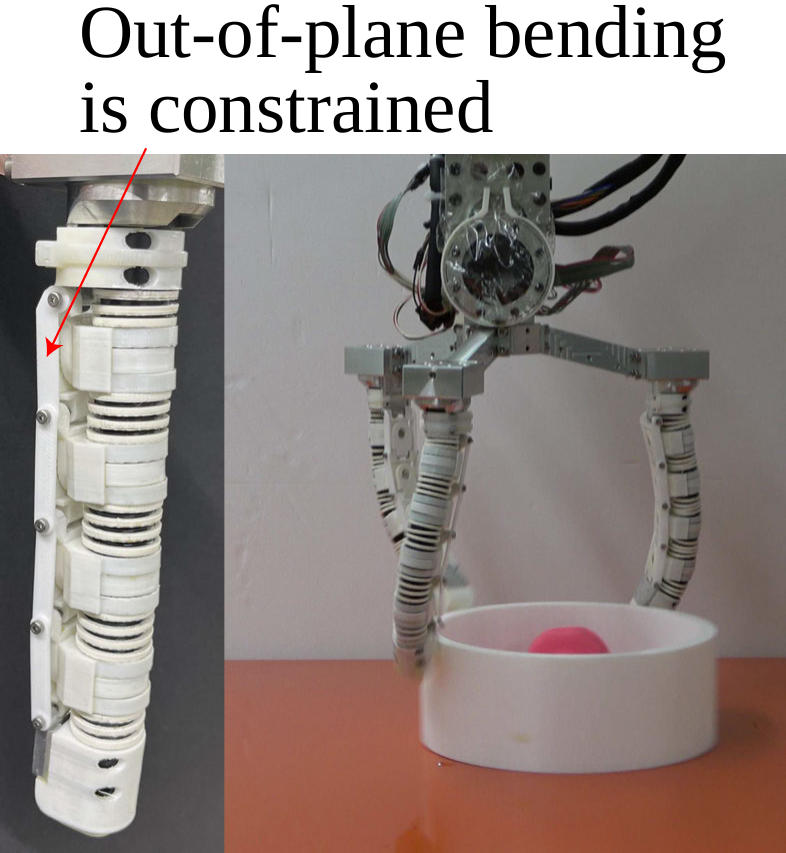}
    }
  \caption{Grasping experiments using the wrapping method: (a) cylindrical box (containing a 1 kg dumbbell). (b) For comparison, the soft hand with constrained out-of-plane deformation failed to grasp the same cylindrical box.}
  \label{fig:grab_various_objects}
  \vspace{-1em}
\end{figure}
%%%%%%%%%%%%%%%%%%%%%%%%%%%%%%%%%%%%%%%%%%%%%%%%

Next, we demonstrated the grasping of various objects using the wrapping motion, including a cylindrical box with 1 kg dumbbell (total 1.26 kg, $\phi 200 \times 100$ mm) and a medicine ball (3 kg, $\phi 150$ mm).
In the following demonstrations, we set the target value of pressure control to 1.5 MPa.
As illustrated in \reffig{fig:grab_various_objects} and accompanying video, the developed fingers can grasp these objects by wrapping.
In particular, \reffig{fig:grab_various_objects_ball} shows that the developed hand could hold a weight of 3 kg.
This is heavier than $m=$1.5 kg assumed in Section \ref{eq:moment_required} because The grasping force was directed upward compared to the scenario considered in \reffig{fig:required_function}.

For comparison, we conducted another experiment to grasp a cylindrical box using a hand with constrained out-of-plane deformation, as shown in \reffig{fig:1DOF_constraint_cylinder}.
However, this experiment failed because sufficient friction was not obtained in a point contact between each finger tip and the lateral surface of the object.
This result demonstrates the validity of the grasping configuration by wrapping, as shown in \reffig{fig:grab_various_objects_cylinder}.

\subsection{Densely Contained Cabbages\label{sect:experiment_cabbages}}
\begin{figure}[t]
  \centering
    \subfloat[\label{fig:cabbage_grabbing_examples_a}]{
        \includegraphics[width=0.9\hsize]{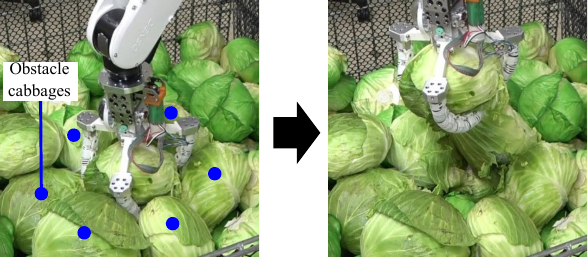}
    }
    \\
    \subfloat[\label{fig:cabbage_grabbing_examples_b}]{
        \includegraphics[width=0.3\hsize]{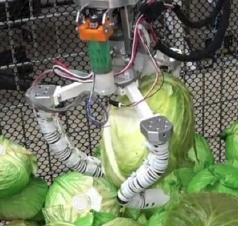}
    }
    \subfloat[\label{fig:cabbage_grabbing_examples_c}]{
        \includegraphics[width=0.3\hsize]{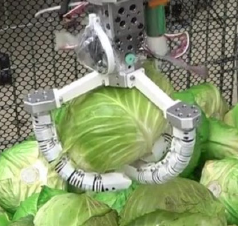}
    }
    \subfloat[\label{fig:cabbage_grabbing_examples_d}]{
        \includegraphics[width=0.3\hsize]{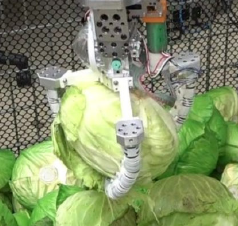}
    }
  \caption{Various scenarios in which the developed hand grasped and picked up a cabbage. In particular, \reffig{fig:cabbage_grabbing_examples_b} shows a scenario in which the finger bent like an S-shape because the inserting this finger was blocked by cabbages.}
  \label{fig:cabbage_grabbing_examples}
\end{figure}

A potential application of the developed hand is the task of picking a cabbage in a vegetable factory \cite{azami2023development,ishibashi2023compact}.
Figure \ref{fig:cabbage_grabbing_examples_a} shows an experimental setup replicating the scenario seen in a vegetable factory, in which a number of cabbages are densely packed in boxes and transported.
In an actual factory, each box contains 100 -- 150 cabbages, and the diameter and weight of each cabbage are approximately 20 -- 30 cm and 2 -- 3 kg, respectively.
In the experiment, the developed hand attempted to grasp an actual cabbage surrounded by other cabbages, as indicated by the blue points in Fig. \ref{fig:cabbage_grabbing_examples_a}.
To replicate this scenario, the cabbages were also surrounded by {\it dummy} cabbages made of styrene foam.

%%%%%%%%%%%%%%%%%%%%%%%%%%%%%%%%%%%%%%%%%%%%%%%%
%%%%%%%%%%%%%%%%%%%%%%%%%%%%%%%%%%%%%%%%%%%%%%%%

Figures \ref{fig:cabbage_grabbing_examples_b} -- \ref{fig:cabbage_grabbing_examples_d} show various scenarios in which the developed hand picks up the cabbage.
Even though there was a case that it was difficult to insert fingers straight in gaps among cabbages, the flexibility of the finger in the out-of-plane direction was effectively utilized in this task.
In particular, \reffig{fig:cabbage_grabbing_examples_b} shows a scenario in which the finger was adaptively bent like to form an S-shape as it was blocked by cabbages.
Nevertheless, the target cabbage was successfully grasped because of its variable stiffness mechanism.

%%%%%%%%%%%%%%%%%%%%%%%%%%%%%%%%%%%%%%%%%%%%%%%%
\begin{figure}[t]
  \centering
    \subfloat[\label{fig:comparison_experiment_old_1}]{
        \includegraphics[width=0.35\hsize]{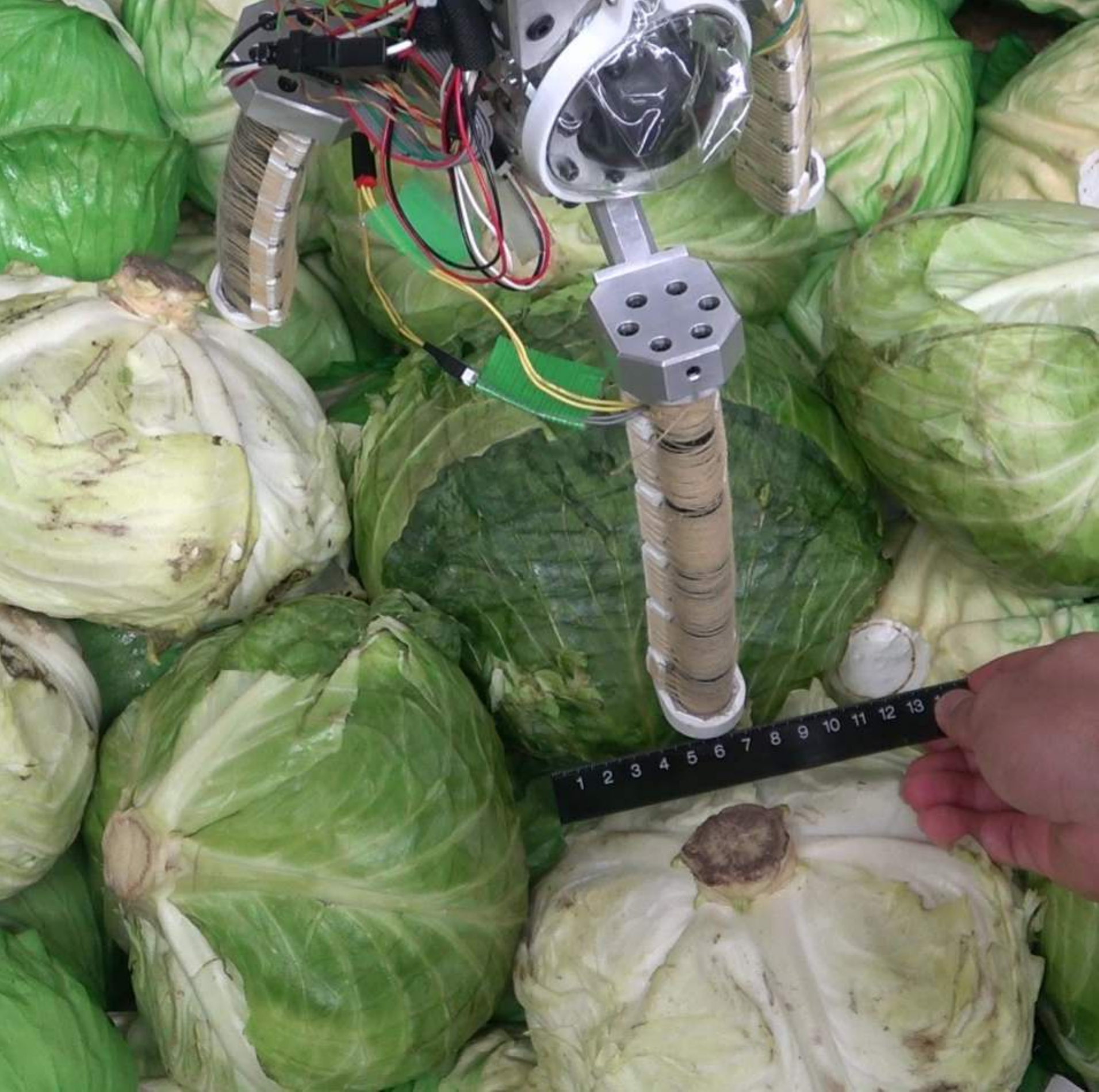}
    }
    \subfloat[\label{fig:comparison_experiment_old_3}]{
        \includegraphics[width=0.35\hsize]{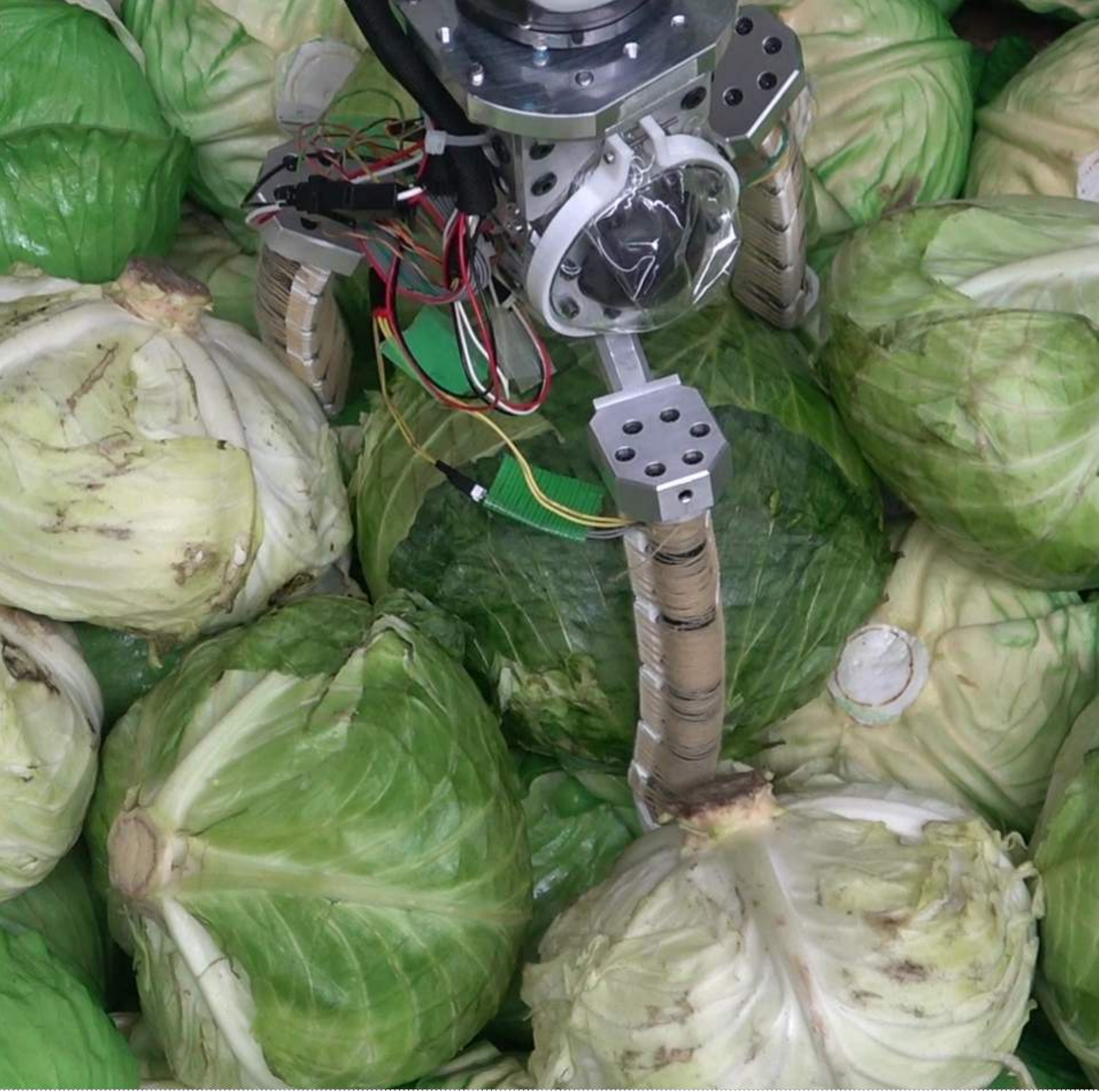}
    }
    \\
    \subfloat[\label{fig:comparison_experiment_new_1}]{
        \includegraphics[width=0.35\hsize]{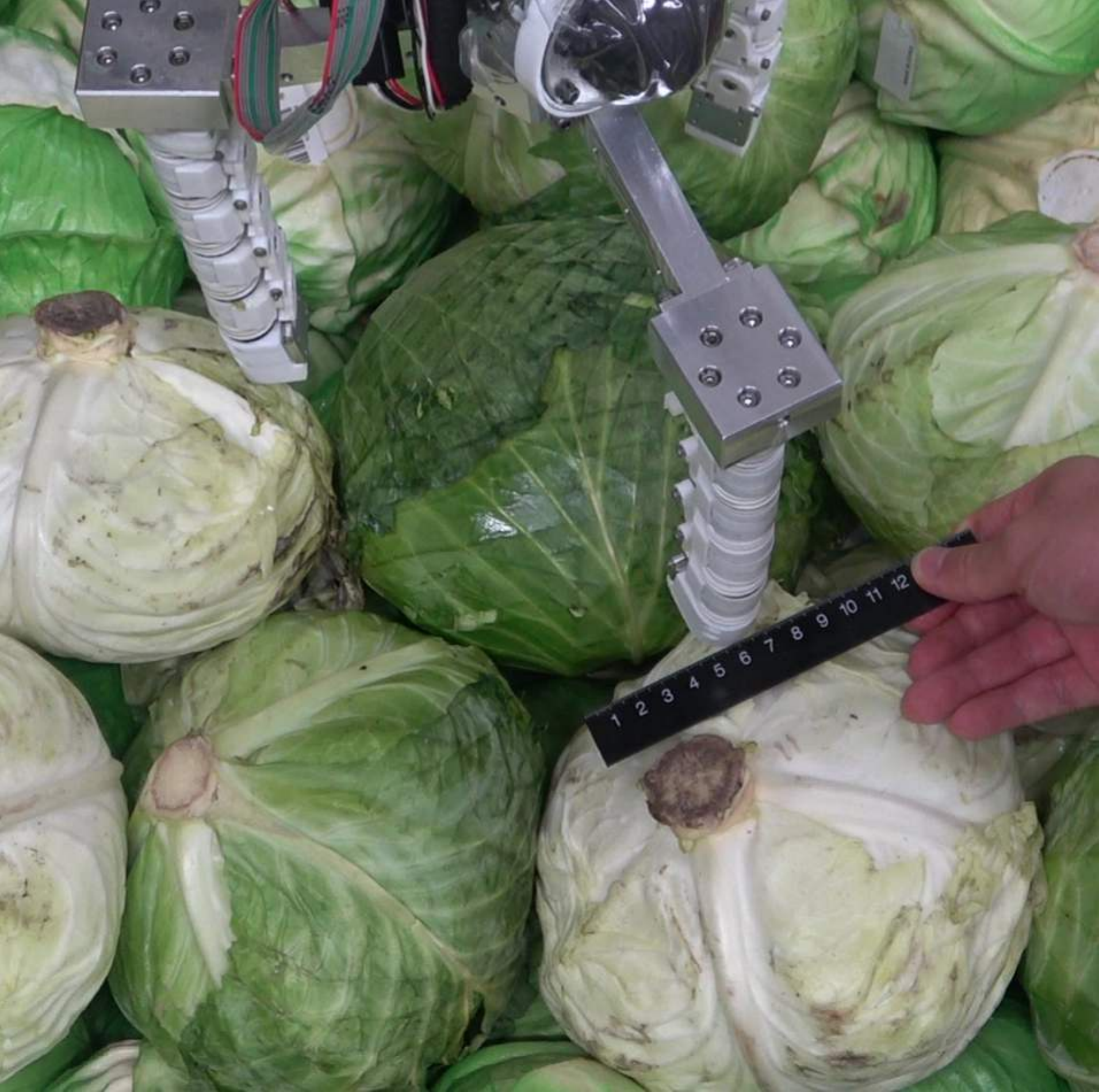}
    }
    \subfloat[\label{fig:comparison_experiment_new_3}]{
        \includegraphics[width=0.35\hsize]{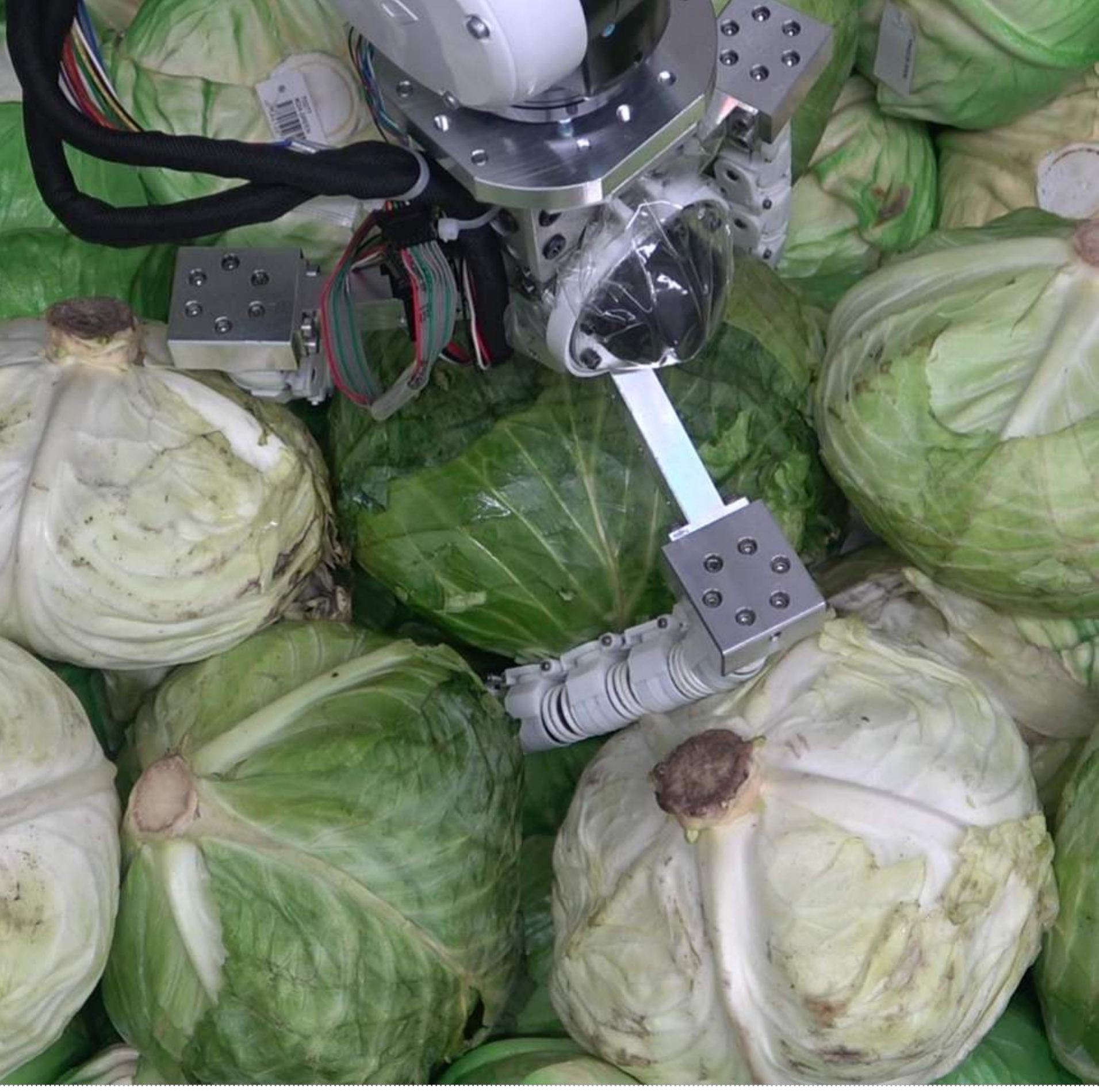}
    }
  \caption{Comparison of the performance between the previously developed hand in \cite{azami2023development} ((a) and (b)) and the developed hand ((c) and (d)).}
  \label{fig:comparison_experiment}
  \vspace{-1em}
\end{figure}
%%%%%%%%%%%%%%%%%%%%%%%%%%%%%%%%%%%%%%%%%%%%%%%%
We also compared the performance of the developed hand with that of the hand developed from \cite{azami2023development}, in which out-of-plane motion was constrained.
In this experiment, we moved the center of hand such that there was some distance between a finger and a gap where the finger can be inserted.
Figure \ref{fig:comparison_experiment} shows experimental scenarios, in which the distance was 6 cm.
 \reffig{fig:comparison_experiment_old_1} and \reffig{fig:comparison_experiment_old_3} show those of the previously developed hand whereas \reffig{fig:comparison_experiment_new_1} and \reffig{fig:comparison_experiment_new_3} show those of the hand developed in this study.
As shown in \reffig{fig:comparison_experiment_old_3}, the previously-developed hand pushed and injured the neighboring cabbage.
This situation would occur even if we could precisely estimate the position of the gap by image processing because gaps are not always aligned in reality.
As shown in \reffig{fig:comparison_experiment_new_3}, however, the developed hand successfully inserted its fingers into the gap, leveraging its out-of-plane flexibility. 
\section{Discussion\label{sect:discussion}}
While we considered out-of-plane bending around the $x$-axis, as shown in \reffig{fig:adaptive_twist_concept}, additional bending can be added around the $z$-axis --- the center axis of the rod structure.
Such a bidirectional twist bending would increase contact area, as shown in \reffig{fig:wrapping_grasp_concept}.
However, a challenge in mechanical design is achieving a compact variable stiffness mechanism, since directly extending the proposed structure to a bidirectional one would make the mechanism larger.

A limitation of the current system is the absence of a bending sensor in the developed soft finger.
Because the purpose of this study is to develop an adaptively deformable soft finger, we focus on the variable stiffness mechanism but not on a feedback control using bending sensors.
Future work includes implementing bending sensors and integrating a feedback control.

In the experiments, the reference profile of the pressure control was manually sent to the hand from a host PC.
In \cite{Ishibashi2025AR}, we reported an autonomous system that integrated the pressure control with bending angle feedback and robotic manipulator motion planning based on image recognition.
Since the control system used in this study is the same as that used in \cite{Ishibashi2025AR}, it is straightforward to integrate the pressure control presented here with such motion planning and vision-based feedback.
\section{Conclusion\label{sect:conclusion}}
In this paper, we developed a soft finger with flexibility for adaptive twist motion in both out-of-plane and in-plane directions, which enabled it to grasp an object by wrapping it with the twisted finger. 
The results are summarized as follows:
\begin{enumerate}
  \item We proposed the variable stiffness mechanism that allowed the finger to generate sufficient stiffness in directions tangential to a contact surface when the finger was pressurized.
    We determined the design parameters of the mechanism based on the FEA, and experimentally verified the FEA results. 
    The experimental results showed a 1.2 Nm maximum moment capacity of the mechanism under 1.5 MPa actuation pressure.
    \item The basic experiment involving the soft hand with three developed fingers showed that the time-constant of the pressure control was 0.3 s regardless of the finger twisting deformation, which implies that the finger exhibited sufficient deformation speed.
    Additionally, we conducted an experiment that involved grasping various objects. The soft hand could hold a 3 kg object, and grasping-by-wrapping was successfully performed, even for an object that the previous soft hand could not grasp.
    \item 
    We conducted the experiment that involved picking up one cabbage from a box of densely contained cabbages. 
    Enabled by the adaptive-twisting deformation, the fingers were easily inserted into gaps among cabbages, and the developed hand could pick up a cabbage whereas the previous hand failed to perform the task.
\end{enumerate}

\bibliographystyle{bib/IEEEtran} 
\bibliography{bib/reference}

%Biography text here without a photo.

%\begin{IEEEbiography}[{\includegraphics[width=1in,height=1.25in,clip,keepaspectratio]{fig1.png}}]{IEEE Publications Technology Team}
%In this paragraph you can place your educational, professional background and research and other interests.\end{IEEEbiography}

\end{document}